%% file: main.tex
\documentclass[fleqn,12pt]{wlscirep}
\usepackage[utf8]{inputenc}
\usepackage[T1]{fontenc}
\usepackage{setspace}
\usepackage{float}

\usepackage{multicol, multirow}
\usepackage{subcaption}
\usepackage{xcolor,colortbl}
\usepackage{xr}
\usepackage{amsmath}
\usepackage{threeparttable}

\title{ Enabling Collaborative Clinical Diagnosis of Infectious Keratitis by Integrating Expert Knowledge and Interpretable Data-driven Intelligence}
\definecolor{Gray}{gray}{0.85}
\definecolor{Dark}{gray}{0.75}

\author[1,2,+]{Zhengqing Fang}
\author[1,6,+]{Shuowen Zhou}
\author[1,2]{Zhouhang Yuan}
\author[1,2]{Yuxuan Si}
\author[2]{Mengze Li}
\author[4]{Jinxu Li}
\author[1,6]{Yesheng Xu}
\author[1,6]{Wenjia Xie}
\author[2,6]{Kun Kuang}
\author[5, *]{Yingming Li}
\author[2,6,*]{Fei Wu}
\author[1,6,*]{Yu-Feng Yao}

\affil[1]{Department of Ophthalmology, Sir Run Run Shaw Hospital, Zhejiang University School of Medicine, Hangzhou, China}
\affil[2]{College of Computer Science and Technology, Zhejiang University, Hangzhou, China }
\affil[3]{School of Public Health, Zhejiang University, Hangzhou, China }
\affil[4]{School of Software Technology, Zhejiang University, Hangzhou, China }
\affil[5]{College of Information Science and Electronic Engineering, Zhejiang University, Hangzhou, China }
\affil[6]{Key Laboratory for Corneal Diseases Research of Zhejiang Province, Hangzhou, China }
\affil[*]{corresponding yaoyf@zju.edu.cn}
\affil[+]{these authors contributed equally to this work}

\begin{abstract}
\footnotesize
\input{_abstract2.tex}

\end{abstract}
\begin{document}
\flushbottom
\maketitle
\thispagestyle{empty}

\label{sec1}
\section*{Introduction}

\begin{figure*}
    \centering
    \includegraphics[width=0.9\textwidth]{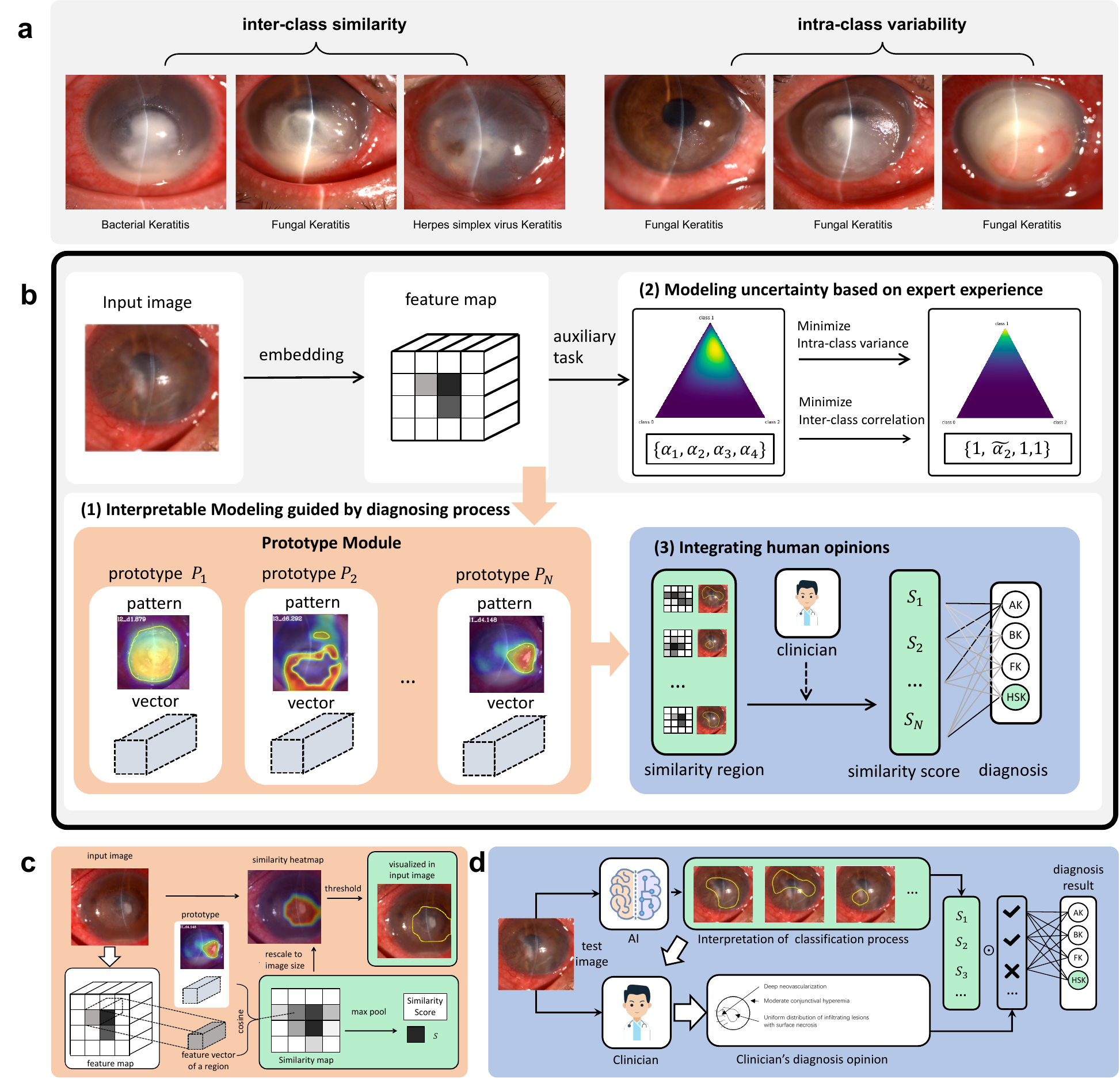}
    \caption{
    \textbf{(a)} Different subclasses of infectious keratitis share similarities (left) but the images from the same subclass may exhibit heterogeneity (right).
    \textbf{(b)} The overview of knowledge-guided data-driven model (KGDM), where the classification depends on similarities between features of the input image and the prototypical parts. The latter are automatically learned from training data with the penalty of minimizing intra-class variance and inter-class correlation based on expert experience. Each learned prototype is embedded into a vector and can be visualized as patterns in historical cases. Based on the visualized interpretation, Humans can incorporate their diagnosis opinion into the classification process by intervention in the weighted sum of similarities.
    \textbf{(c)} Detailed description of how to visualize a learned prototype by its embedding vector on a test image. The high similarity area is circled within a yellow contour and the corresponding maximum similarity is used to classify.
    \textbf{(d)} Detailed description of human-AI complementary diagnosis. Clinicians could take a look at the interpretation of the classification process as a reference and give an AI-aided diagnosis opinion. Combined with the clinician's opinion of the current case patterns, the original reasoning process of KGDM can also be modified to neglect incorrect prototypes.}
\label{fig:KGDM}
\end{figure*}

\input{_intro_npj.tex}

\section*{Method}

\subsection*{Data collection}

\begin{figure*}
    \includegraphics[width=0.9\linewidth]{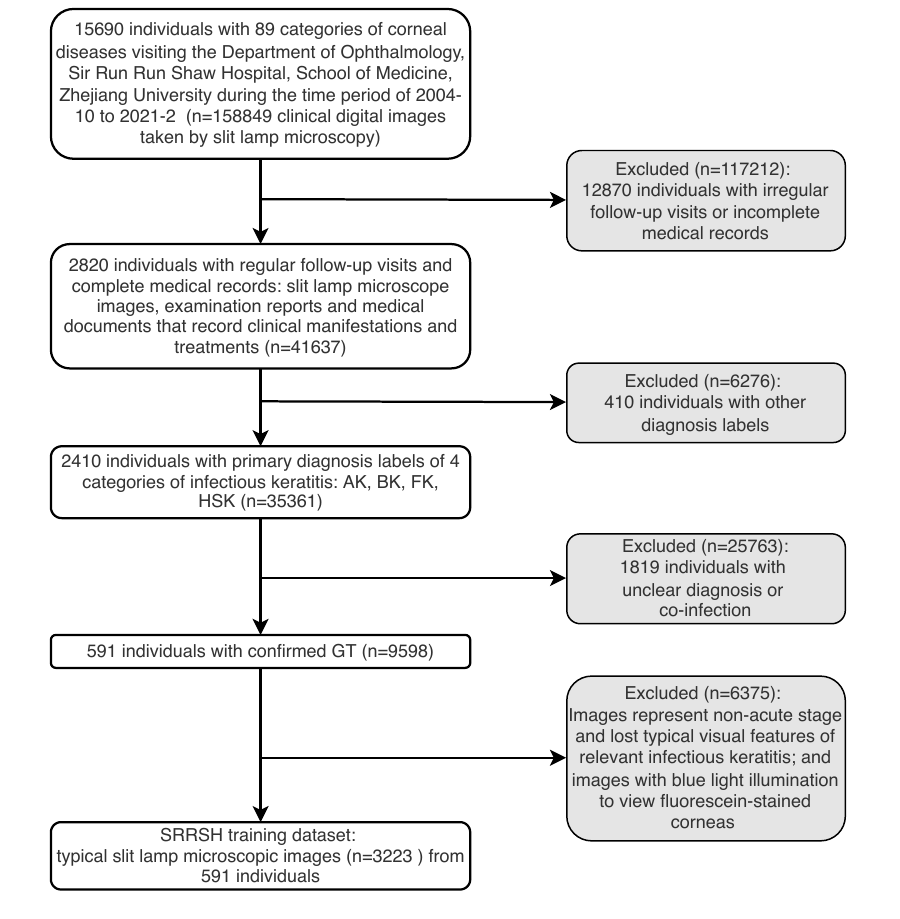}
    \caption{\textbf{Inclusion and exclusion criteria of the SRRT.}
The visiting time, diagnosis, medical records, and slit lamp microscopic images of each case were rechecked by ophthalmologists via a standard process to include and exclude the images in the SRRT. We strictly selected 3223 high-quality images from 35361 images to train the case-based interpretation model.}
    \label{fig:inclusion and exclusion}
\end{figure*}

Infectious keratitis (IK) includes acanthamoeba keratitis (AK), bacterial keratitis (BK), fungal keratitis (FK), herpes simplex virus keratitis (HSK), and keratitis caused by other pathogens. 
The training dataset (SRRT) for this study was retrospectively collected from 129853 clinical digital images of 89 categories of corneal diseases.
These clinical images were taken by slit lamp microscopy during the time period of May 2004 to  in the Department of Ophthalmology, Sir Run Run Shaw Hospital, School of Medicine, Zhejiang University (SRRSH).
Two types of equipment were used to take the images, SL 130 (Carl Zeiss Meditec AG, Germany) with a resolution of 1024 $\times$ 768 pixels; and Topcon slit lamp microscope (TOPCON Corporation, Japan), with digital camera Unit DC-1 offering an image resolution of 1740 $\times$ 1536 pixels or 2048 $\times$ 1536 pixels. 
We restricted the visit time to between January 2014 and January 2022 and excluded the cases with diagnoses other than AK, BK, FK, and HSK.
The ground truth of diagnosis was corroborated by ophthalmologists of SRRSH with at least two pieces of the following evidence:
(1) the responsiveness to appropriate drug treatment, (2) the detection of specific pathogens, and (3) typical clinical characteristics.
Cases with an unclear diagnosis or co-infection, where more than one pathogen was involved in a single case, were excluded.

For the prospective validation dataset (SRRPV), we prospectively enrolled patients who visited the Department of Ophthalmology, Sir Run Run Shaw Hospital, School of Medicine, Zhejiang University for \emph{``blurred vision with eye irritation or eye pain''} between May 2021 and June 2022. The ground truth labels were annotated as the same as the training dataset. 
The external test dataset (XS) was collected from March 2019 to 2022 in the Department of Ophthalmology, Xiasha Branch of Sir Run Run Shaw Hospital, School of Medicine, Zhejiang University with the same ground truth establishing criteria as the SRRSH training dataset.
We also collected images from the publicly available dataset to evaluate robustness, by searching Pubmed for slit lamp microscopic images using the keywords \emph{``keratitis''}, \emph{``amoeba keratitis''}, \emph{``bacterial keratitis''}, \emph{``fungal keratitis''} and \emph{``herpes simplex virus stromal keratitis''}. Images were reviewed by two ophthalmologists in our dataset for reliability. A subset of the Human-XAI collaboration test dataset was collected in the same way as the SRRSH prospective validation dataset.

\subsection*{Design of Knowledge Guided Data-driven Model}
\label{sec:KGDM}
The overall framework of KGDM is shown in Figure\ref{fig:KGDM}. (1) Interpretable modeling motivated by the case-based diagnosing process. A prototype layer was introduced in the KGDM to identify the key parts of the images which resembled the learned prototypical parts of certain classes. This assisted clinicians in understanding the particular diagnostic patterns of KGDM by referring to the prototypical parts. (2) Modeling uncertainty based on evidence accumulating. An evidential deep learning task was introduced in KGDM, where it sought to predict Dirichlet distribution parameters instead of direct class probabilities, differing from vanilla classification models \cite{Simonyan2015, He2016, Huang2017a}. This task aimed to jointly optimize the prediction error, the variance, and the inter-class correlation of the predictive distribution, which reduced the influence of confounding patterns during network training.
(3) Integrating human expert opinions via human-AI interaction. The auto-learned diagnostic patterns from large-scale datasets can be regarded as AI-based diagnostic markers, providing clinicians with rich pieces of evidence for making diagnoses, which can be viewed as a tool to complement their diagnostic experiences. An interactive interface was designed for clinicians to conduct optional test-stage interventions \cite{Koh2020} to further improve the model performance.

The slit-lamp images were scaled to 384 $\times$ 384 pixels, with random flip, rotation, and brightness augmentation.
The backbone used to embed the images was optional, we used ResNet50 for the balance of computing efficiency and performance.
The size of the feature map we obtained was 12 $\times$ 12 $\times$ 2048.
For prototype layers, we set 10 prototypes for each class and randomly initialized it with the size of 40 $\times$ 2048,
where each prototype was represented with a 2048-dimensional vector.
After training, we visualized each prototype by point-wise similarity and drew a yellow contour around the high-similarity area.
The training images with the highest similarities were chosen to represent the manifestation prototype.
The visualization process is shown in Figure~\ref{fig:KGDM}(c).

\subsection*{Comparison of classification performance}
KGDM was compared with purely data-driven models like VGG16 \cite{Simonyan2015}, DenseNet121\cite{Huang2017a}, ResNet50 \cite{He2016} and Vision transformer (ViT \cite{Dosovitskiy2021a}).
The deep models were trained on the SRRST dataset with a 5-fold cross-validation strategy and evaluated on SRRPV, XS, and PUB datasets.
All test patients and their images in three datasets were not previously seen by the AI models.
The same data augmentation strategy and training hyperparameters were employed for all models.
Given a test image, the models output a probability for each of the 4 classes.
Predicted results were post-processed by an uncertainty threshold of 0.7 measured by normalized entropy \cite{Chua2022}.
The AI models were evaluated on the following metrics: area under the receiver operating characteristic (AUROC), Cohen's kappa statistic (Cohen's Kappa), sensitivity, and positive predictive value (PPV).
The average performance with a 95\% confidence interval (CI) was reported for each metric.
The class-wise area under the curve (AUC) plots were provided for all three datasets as shown in Figure \ref{fig: cls_wise}a.
To demonstrate the internal change in KGDM, we visualized the  feature distribution of SRRT and SRRPV via 2D t-SNE visualizations. 

\subsection*{Interpretation of learned manifestation prototype and its evaluation}
In KGDM, the learned prototypes of keratitis are the basic units for interpretation, as their representation vectors can provide the corresponding visualization for human comprehension. We set 10 prototypes for each of the four target subclasses of infectious keratitis (including AK, BK, FK, and HSK), which were indexed from $P_0$ to $P_{39}$. 
To visualize each prototype on an image, a cosine similarity map can be calculated between its feature map and each learned prototype vector. Scaled to the same size as the input image, it became a heatmap to show which parts of the image were similar to the learned prototype. The high similarity area was circled within a yellow contour, as shown in Figure~\ref{fig:KGDM}c. Further, based on the visualization of prototypes, we can visualize the reasoning process of a sample for interpretation.
Since the classification is performed by the weighted sum of similarities to different prototypes, clinicians are concerned about what the most similar prototypes are and how they contribute to the final classification score.
The similar regions of the top contribution prototypes are visualized on the test sample together with the closest images of the same prototype in the training dataset. 

The learned protoypes can serve as AI-based imaging biomarkers for diagnosing infectious keratitis. To quantitatively investigate their effectiveness as individual diagnostic markers, their diagnostic odd ratios (DOR) are computed on the prospective SRRPV dataset.
This metric is calculated as the ratio of the odds of positive test results in the target class to that in the non-target class. The higher the DOR score indicates the more positive correlation between the prototype and the target class. The DOR scores of each prototype and each target class are shown in Figure~\ref{fig:interpretation}b. 


\subsection*{Human-XAI collaboration test}
\label{sec:collaborated}
To assess the feasibility of KGDM as a tool to aid clinicians in diagnosing infectious keratitis in real-world clinical environments, 12 ophthalmologists were recruited in a simulated clinical diagnostic test to evaluate the collaborative performance of combining clinicians and KGDM.
The test was conducted in two consecutive steps: (1) the ophthalmologists made their own prediction without KGDM prediction, and (2) they were asked to make a new prediction using the interactive interface shown in Figure~\ref{fig:hmc}.
Figure \ref{fig:KGDM+human}a illustrates the collaborating workflow.
With the visualized classification process, the testers could subjectively assess the similarity and drop out the prototypes, then KGDM discarded their corresponding contributions.
They would continue interaction until they reached a consensus for each sample.
The testers had been sufficiently educated on the KGDM and the interactive interface before the test.
The sensitivity (ss) and specificity (sp) before and after using KGDM were reported.
Fleiss' kappa ($\kappa^F$), which assesses the reliability of agreement between a fixed number of raters when classifying items, was used to measure the agreement of ophthalmologists.
The performance was analyzed in two subgroups according to testers' years of experience, denoted as Junior (< 5 years, n=6) and Senior (> 5 years, n=6), as the diagnostic performance is divergent across different ophthalmologists \cite{Xu2021}.
For Junior groups, we investigated the performance gained by collaborating with KGDM.
For Senior groups, we were interested in how much the AI increased from ophthalmologists' correction.
The results are shown in Table \ref{tab:KGDM+human}, Figure \ref{fig:KGDM+human}b, and \ref{fig:KGDM+human}c.

\subsection*{Statistical analysis}
All statistical analyses were performed using Python (version 3.10.1, Python Software Foundation, State of Delaware, USA).
The graphs and charts were created based on Matplotlib 3.3.2 and Seaborn 0.12.2.
For each of the experiments, the following parameters were calculated on the test set: ROC-AUC, cohenKappa, sensitivity, and PPV.
Among 6 different random seeds results of 5-fold cross-validation, we reported the mean and 95\% confidence interval for each metric.
For the interpretation of AI-based biomarkers(manifestation prototypes), we calculated the diagnostic odds ratio and p-value using Scipy statistical function V1.11.1.
Differences in the variation of the diagnostic sensitivity, specificity, and accuracy after collaborating with KGDM between the Junior/ Senior groups were analyzed using the paired Wilcoxon test. The significance level for all the tests was set to 0.05.
For the clinical diagnosis experiments, Fleiss' Kappa has been used to calculate inter-reader agreement between the 12 ophthalmologists.

\input{_results.tex}

\input{_discussion.tex}
\section*{Code availability}
Code of KGDM and the Human-XAI collaboration test analysis is available at \href{https://github.com/ZJU-AIEYE/KGDM.git}{https://github.com/ZJU-AIEYE/KGDM.git}.

\section*{Data availability}
We provided download links for each image in the PUB dataset in Table \ref{tab:pub_download}, but the images in SRRT, SRRPV, and XS datasets are not publicly available because they contain confidential information that may compromise patient privacy as well as the ethical or regulatory policies of our institution. Data will be made available on reasonable request, for non-commercial research purposes, to those who contact Y.F Yao. (yaoyf@zju.edu.cn). Data usage agreements may be required. Source Data are provided in this paper.

\section*{Acknowledgements }
This project was supported by the National Key R\&D Program of China (2022ZD0118003), the National Natural Science Foundation of China (Grant No. U20A20387), the Health Commission of Zhejiang Province (Grant No. 2019ZD040).

\section*{Author contributions statement}
Z.F., Z.Y. designed the algorithm.
J.W., S.Z., Y.S., W.X., Y.X built the dataset.
Z.F., Z.H, Y.L., K.K. conceived the experiment(s),
Z.F., S.Z., J.L. conducted the experiment(s)
Z.F. and S.Z. analyzed the results.
Z.F, Y.L, S.Z. wrote the manuscript.
All authors reviewed the manuscript. 




\bibliography{mybib}

\input{_methods}
\input{supplementary}


\end{document}

%% file: _abstract2.tex
Although data-driven artificial intelligence (AI) in medical image diagnosis has shown impressive performance in silico, the lack of interpretability makes it difficult to incorporate the "black box" into clinicians' workflows. To make the diagnostic patterns learned from data understandable by clinicians, we develop an interpretable model, knowledge-guided diagnosis model (KGDM), that provides a visualized reasoning process containing AI-based biomarkers and retrieved cases that with the same diagnostic patterns. KGDM also embraced clinicians' prompts into the interpreted reasoning through human-AI interaction, leading to potentially enhanced safety and more accurate predictions. This study investigates the performance, interpretability, and clinical utility of KGDM in the diagnosis of infectious keratitis (IK), which is the leading cause of corneal blindness. The classification performance of KGDM is evaluated on a prospective validation dataset, an external testing dataset, and an publicly available testing dataset. The diagnostic odds ratios (DOR) of the interpreted AI-based biomarkers are effective, ranging from 3.011 to 35.233 and exhibit consistent diagnostic patterns with clinic experience. Moreover, a human-AI collaborative diagnosis test is conducted and the participants with collaboration achieved a performance exceeding that of both humans and AI. By synergistically integrating interpretability and interaction, this study facilitates the convergence of clinicians' expertise and data-driven intelligence. The promotion of inexperienced ophthalmologists with the aid of AI-based biomarkers, as well as increased AI prediction by intervention from experienced ones, demonstrate a promising diagnostic paradigm for infectious keratitis using KGDM, which holds the potential for extension to other diseases where experienced medical practitioners are limited and the safety of AI is concerned.

%% file: _intro_npj.tex
Artificial intelligence (AI) has catalyzed tremendous progress in medical image diagnosis \cite{McKinney2020, Tschandl2020, Bai2021}.
Neural networks, typically consisting of many layers connected via many nonlinear intertwined relations, have the ability to learn complex diagnostic patterns from large-scale data.
Despite indications that some of them match the accuracy of human clinicians within in silico studies \cite{Liu2019b}, it is still an open question that how much these models can facilitate clinician decision-making performance \cite{Vasey2021}.
We called them \emph{``black-box AI''} because even if one is to inspect all model layers and describe their relations, it is unfeasible to fully comprehend how the neural network came to its decision \cite{VanDerVelden2022}.
The disease manifestations, \emph{i.e.} features, are implicitly mapped from input images to classification results in neural networks.
It mismatches the real-world diagnosis where many factors are considered by clinicians according to different purpose.
For example, in diagnosing infectious keratitis, clinicians not only distinguish if an eye is keratitis but also need to discriminate different subtypes. Some manifestations might be confounding and need to be excluded from diagnosis, which implies what the clinicians need is more than an image classification result, but an intervenable reasoning process showing how the data-driven intelligence comes to its decision.

As one of the primary causes of corneal blindness \cite{Ting2021, Cabrera-Aguas2022}, infectious keratitis (IK) is caused by bacteria, fungi, parasites, viruses, and other pathogen infections \cite{Clemens2017, Juarez2018, Dalmon2012}. Prompt identification of the etiology is critical for ensuring effective medical treatment and preventing deleterious outcomes \cite{Redd2022b, Ting2022}. Empirical therapy has to be conducted according to observed manifestations as the golden standard diagnosis result is time-consuming, invasive, and sometimes unavailable \cite{Dalmon2012}. 
Ophthalmologists rely on their experience of observing slit-lamp photographs, from which they can summarize and memorize diagnostic patterns. However, images within the same subclass have a considerable variance, whereas images from different subclasses may exhibit notable similarities (shown in Figure \ref{fig:KGDM}a). It results in divergent diagnostic performance among ophthalmologists \cite{Xu2021,Redd2022b}, and less experienced individuals often achieve less satisfactory performance.
To avoid false diagnoses, inexperienced ophthalmologists seek diagnostic opinions from their experienced colleagues whenever possible.
Considering the limited availability of experienced experts relative to the large patient population, utilizing data-driven AI to interpret diagnostic patterns could be a promising solution if it provides comparable performance to those of experienced experts and reliable interpretation to clinicians. While there already exists several studies try to classify infectious keratitis images~\cite{Ghosh2022, Redd2022}, the majority of them have relied on black-box deep learning models such as VGG~\cite{Simonyan2015}, ResNet~\cite{He2016}, and DenseNet~\cite{Huang2017a}. To date, how to design an interpretable AI model to diagnose infectious keratitis is still an open problem. 

In this study, we focus on the diagnosis of infectious keratitis and developed a self-explainable classification model that can communicate with ophthalmologists. Technically, the proposed method is inherit from ProtoPNet~\cite{Chen2019} and evidential deep learning (EDL)~\cite{Sensoy2018}.
ProtoPNet is a deep network architecture that reasons similarly to how humans dissect images by finding prototypical parts and utilizing similarities to the prototypes to make a final classification. It integrates the capability of deep learning and the interpretability of case-based reasoning. The learned prototypical patterns can be viewed as AI-based biomarkers.
Additionally, EDL views the learning process as an evidence-acquisition process, whose loss function guarantees that model predictions are derived from observed evidence of the target class, rather than relying on the absence of evidence from other classes.
This enables the learned model to identify the evidence that minimizes uncertainty and say \emph{``I do not know''} when encountering images not belonging to any of the candidate classes.
As shown in Figure~\ref{fig:KGDM}b, we integrate the prototype layer into our model architecture and use EDL to fortify its robustness.
Furthermore, we devise an interactive interface to visualize the reasoning process, facilitating manual adjustments to the prediction results after considering clinical factors.
Overall, our approach combines the case-based diagnosis method, evidence accumulation, and communication interface within a data-driven AI framework, mirroring the diagnostic approach of clinicians while seamlessly integrating their individual knowledge. We aptly refer to this model as the \emph{``knowledge-guided diagnosis model(KGDM)''}.
We evaluate the classification performance of KGDM on a prospective validation dataset, an external testing dataset, and an publicly available testing dataset. The diagnostic odds ratios (DOR) of the interpreted AI-based biomarkers are also assessed on the prospectively collected dataset. Moreover, we conduct a human-AI collaborative diagnosis test and analyzed the performance of 12 stand-alone ophthalmologists with and without interacting with KGDM. The results demonstrate that KGDM can achieve comparable performance to experienced ophthalmologists and significantly improve the performance of junior ophthalmologists.

%% file: _results.tex
\section*{Results}
\subsection*{Data characteristics and dataset collection}
In our study, a total of 5 datasets have been collected for the training of AI models, prospective validation in clinics, testing in external medical centers, and Human-XAI collaboration test.
The clinical characteristics of the participants are shown in Table \ref{tab:data}.
(1) The SRRSH training dataset (SRRT, n=591) contained 1386 HSK images, 527 BK images, 977 FK images, and 333 AK images(The filtrating procedure is shown in Figure~\ref{fig:inclusion and exclusion}). The  mean age of the patients in this dataset was 55.01 (standard deviation [SD], 15.37) years and the proportion of female patients was 37.93\%;
(2) In the SRRSH prospective validation dataset (SRRPV, n=161) containing 88 HSK images, 92 BK images, 138 FK images, and 153 AK images, the patients' mean age was 48.89 (standard deviation [SD], 15.54) years and the proportion of female patients was 50.93\%;
(3) In the XS hospital testing dataset (XS, n=120) consisting of 26 images of AK, 31 images of BK, 31 images of FK, and 50 images of HSK, the mean age of patients was 54.15 (standard deviation [SD], 15.50) years and the proportion of female patients was 22.46\%.
(4) The online available publication testing dataset (PUB) contained 15 images of AK, 11 images of BK, 20 images of FK, and 20 images of HSK; 
(5) The Human-XAI collaboration test dataset (HXACT, n=120) consisted of 18 images of AK, 23 images of BK, 25 images of FK, 29 images of HSK, 32 images of other corneal diseases, and 13 images of corneas without significant abnormalities. The mean age of the patients in this group was 51.61 (standard deviation [SD], 15.82) years and the proportion of female patients was 48.18\%.

\begin{table}[thb]
    \centering
    \caption{\textbf{Data characteristic of collected datasets}}
    \begin{threeparttable}
        \resizebox{\textwidth}{!}{
            \begin{tabular}{lccccc}
                \rowcolor{Dark}
                \hline \multicolumn{1}{c}{\text { Dataset }} & \text { SRRT } & \text { SRRPV } & \text { XS } & \text { PUB} & \text { HXACT } \\
                \hline
                \rowcolor{Gray}
                \textbf {Image number } & & & & &\\
                  AK & 333 (10.33\%) & 88 (7.51\%) & 26 (18.84\%) & 15 (22.73\%) & 18 (12.86\%) \\
                 BK  & 527 (16.35\%) & 92 (7.86\%) & 31 (22.46\%) & 11 (16.67\%) & 23 (16.43\%) \\
                 FK  & 977 (30.31\%) & 138 (11.78\%) & 31 (22.46\%) & 20 (30.30\%) & 25 (17.86\%) \\
                 HSK & 1386 (43.00\%) & 153 (13.07\%) & 50 (36.23\%) & 20 (30.30\%) & 29 (20.71\%) \\
                 non-IK\tnote{1} & 0(0.00 \%) & 554 (47.31\%) & 0(0.00 \%) & 0(0.00 \%) & 32 (22.86\%) \\
                 CWNOA\tnote{2}  & 0(0.00 \%) & 146 (12.47\%) & 0(0.00 \%) & 0(0.00 \%) & 13 (9.29\%) \\
                 Total images & 3223 & 1171 & 138 & 66 & 140 \\
                 \rowcolor{Gray}
                \textbf {Patient characteristic} & & & & &\\
                 Total patients  & 591 & 161 & 120 & NA & 120 \\
                 Patient age (mean±std)  & 55.01±15.37 & 48.89±15.54 & 54.15±15.50 & NA & 51.61±15.82 \\
                 Female patients (\%)  & 37.39\% & 50.93\% & 22.46\% & NA & 48.18\% \\
                \hline
            \end{tabular}
            }
    \begin{tablenotes}
        \footnotesize
        \item[1] Other corneal diseases
        \item[2] Cornea with no obvious abnormality
    \end{tablenotes}
    \end{threeparttable}

\label{tab:data}
\end{table}

\subsection*{Performance comparison between KGDM and \emph{``black box AI''} models}
Table~\ref{tab:performance} (a) presents the diagnostic performance of KGDM on the SRRSH prospective validation dataset. Specifically, KGDM achieves an average area under the curve (AUC) of 0.836 (95\% CI: 0.811, 0.861).
To account for class imbalance, Cohen's kappa ($\kappa$) is used to measure classification performance.
KGDM achieved a $\kappa$ of 0.558 (95\% CI: 0.521, 0.596), which is 36.10\% higher than that of VGG16 (0.410, 95\% CI: 0.305, 0.515).
It also achieves a sensitivity ($ss$) of 62.65\% (95\% CI: 59.88\%, 65.41\%) and positive predictive value ($ppv$) of 70.03\% (95\% CI: 65.20\%, 74.86\%), showing improvements of 19.38\% and 28.41\%, respectively, over those of VGG16.

The diagnostic performance of KGDM on the XS dataset is presented in Table~\ref{tab:performance}(b), achieving an AUC of 0.770 (95\% CI: 0.723, 0.817), $\kappa$ of 0.411 (95\% CI: 0.313, 0.509), $ss$ of 52.83\% (95\% CI: 44.51\%, 61.15\%), and $ppv$ of 61.07\% (95\% CI: 46.71\%, 75.43\%).
The results on PUB dataset have been shown in Table~\ref{tab:performance}(c), where KGDM achieves AUCs of 0.638 (95\% CI: 0.555, 0.721), $\kappa$ of 0.181 (95\% CI: 0.035, 0.328), $ss$ of 37.69\% (95\% CI: 25.89\%, 49.48\%), and $ppv$ of 55.38\% (95\% CI: 42.35\%, 68.42\%).
These results demonstrate that KGDM is more robust than data-driven approaches in various test datasets.

\begin{table}[th]
    \centering
    \caption{\textbf{Performance of KGDM and other deep learning models on the internal prospective validation cohorts and the external test dataset.}}
    \begin{threeparttable}
        \resizebox{\linewidth}{!}{
            \begin{tabular}{lccccc}
                \rowcolor{Dark}
                \hline 
                \multicolumn{1}{c}{\text { Metrics\tnote{1} }} & \text { AUROC (95\% CI)\tnote{2} } & \text { Cohen's Kappa (95\% CI)\tnote{3} } & \text { Sensitivity (95\% CI)\tnote{4} } & \text { PPV (95\% CI)\tnote{5} }   \\
                \hline
                 \rowcolor{Gray}
                \multicolumn{5}{l}{\textbf {(a) Internal Prospective Dataset: SRRPV }}\\
                 ResNet50 &  0.795(0.781-0.809) &  0.160(0.149-0.169) &  59.70\%(57.56\%,61.60\%) &  35.88\%(33.29\%,38.62\%) \\
                 DenseNet121 &  0.764(0.745-0.782) &  0.328(0.206-0.451) &  46.13\%(37.35\%-54.91\%) &  50.86\%(45.64\%-56.08\%) \\
                 VGG16 &  0.789(0.742-0.836) & 0.410(0.305-0.515) & 52.48\%(44.47\%-60.49\%) & 54.46\%(48.70\%-60.21\%) \\
                 ViT-B-16 & 0.791(0.779-0.802) &    0.130(0.121-0.140) &  52.92\%(51.08\%-54.94\%) &  35.25\%(32.74\%-37.20\%) \\
                 KGDM  &   \textbf{0.836(0.811-0.861)} &   \textbf{0.558(0.521-0.596)} &   \textbf{62.65\%(59.88\%-65.41\%)} &   \textbf{70.03\%(65.20\%-74.86\%)} \\
                 \rowcolor{Gray}
                \multicolumn{5}{l}{\textbf {(b) External Retrospective Dataset: XS }}\\
                 ResNet50  &  0.729(0.641-0.817) &  0.283(0.171-0.395) &  44.80\%(37.62\%-51.97\%) &  43.92\%(30.83\%-57.02\%) \\
                 DenseNet121 &  0.721(0.680-0.762) &  0.263(0.184-0.342) &  43.30\%(37.40\%-49.19\%) &  44.30\%(28.81\%-59.80\%) \\
                 VGG16 &  0.769(0.741-0.798) &  0.328(0.248-0.408) &  48.36\%(42.24\%-54.48\%) &  52.43\%(38.32\%-66.55\%) \\
                 ViT-B-16 &  \textbf{0.774(0.757-0.792)} &    0.323(0.288-0.352) &  47.34\%(44.53\%-49.76\%) &  54.19\%(49.40\%-58.62\%) \\
                 KGDM &  0.770(0.723-0.817) & \textbf{ 0.411(0.313-0.509)} &  \textbf{52.83\%(44.51\%-61.15\%)} &  \textbf{61.07\%(46.71\%-75.43\%)} \\
                 \rowcolor{Gray}
               \multicolumn{5}{l}{\textbf {(c) External Retrospective Dataset: PUB }}\\
                 ResNet50 &  0.521(0.420-0.623) &  0.033(-0.182-0.249) &  26.14\%(10.89\%-41.39\%) &  27.68\%(8.31\%-47.05\%) \\
                 DenseNet121 & 0.552(0.443-0.661) &  0.031(-0.184-0.245) &  26.48\%(10.02\%-42.94\%) &  29.40\%(9.65\%-49.16\%) \\
                 VGG16 &  0.574(0.482-0.666) &  0.033(-0.160-0.225) &  26.10\%(12.15\%-40.05\%) &  32.61\%(8.85\%-56.37\%) \\
                 ViT-B-16 &  0.617(0.591-0.644) &    0.126(0.088-0.168) &  32.87\%(29.99\%-35.98\%) &  33.31\%(29.60\%-36.83\%) \\
                 KGDM &  \textbf{0.713(0.683-0.741)} &  \textbf{0.209(0.169-0.245)} &  \textbf{39.57\%(36.76\%-42.24\%)} &  \textbf{49.21\%(42.43\%-56.16\%)}  \\

                \hline
            \end{tabular}
            }
    
    \begin{tablenotes}
        \scriptsize
        \item[1] Reported values are the macro-average of 4 target classes with a 95\% confidence interval. The best values are in bold.
        \item[2] AUROC, the area under the receiver operating characteristic, assesses how well a model can separate positive and negative samples for a given label.
        \item[3] Cohen's kappa statistic can handle both multi-class and imbalanced class problems.
        \item[4] Sensitivity refers to the test's ability to detect disease correctly.
        \item[5] PPV, positive predictive value, refers to how likely it is to be truly positive in the case of a positive test result.
    \end{tablenotes}
    \end{threeparttable}
\label{tab:performance}
\end{table}

\begin{figure*}[th]
    \centering
    \includegraphics[width=\linewidth]{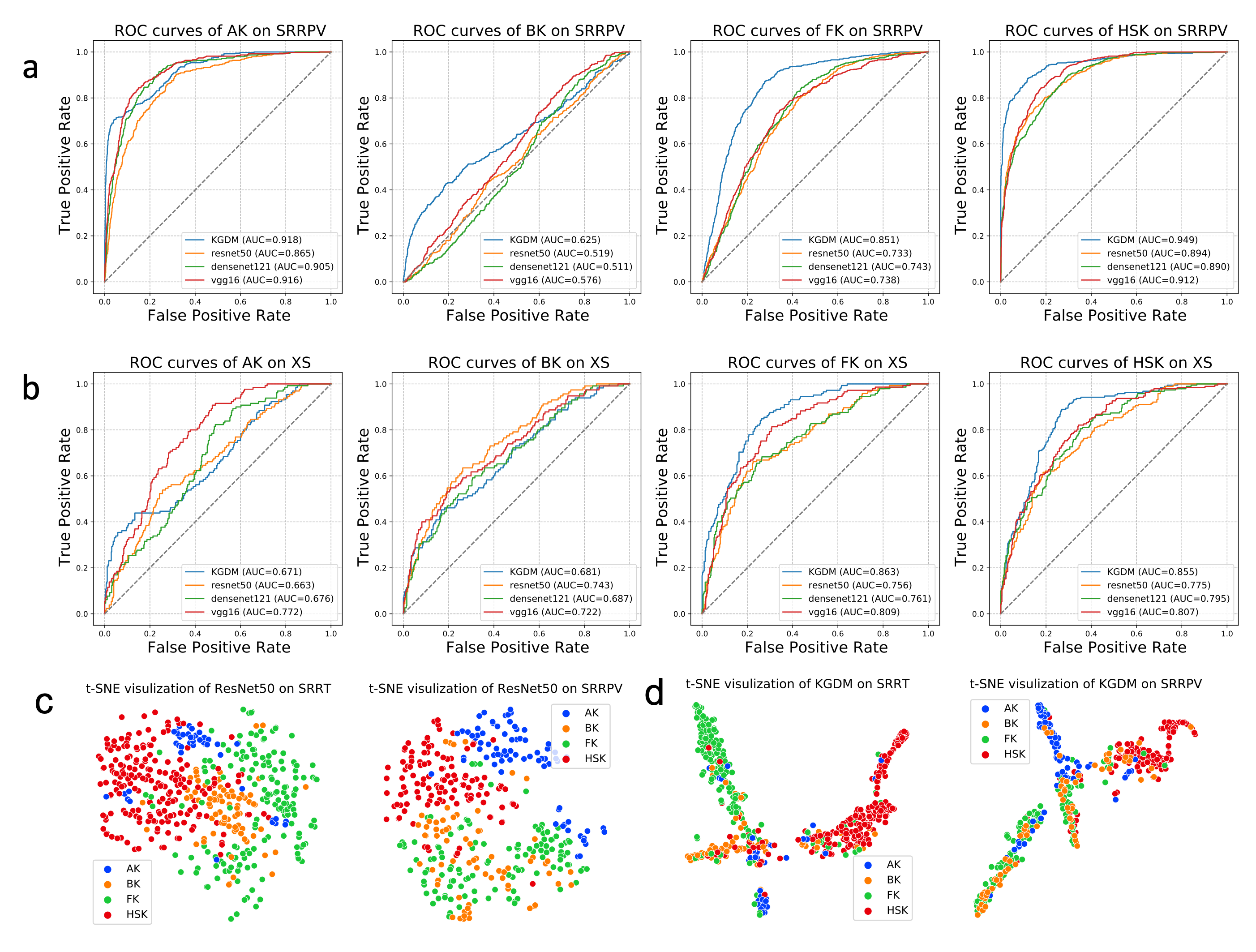}
    \caption{\textbf{Comparison of KGDM and purely data-driven models.}
    The receiver operating characteristic (ROC) curves for diagnosing AK, BK, FK, and HSK on \textbf{(a)} SRRPV and \textbf{(b)} XS. KGDM performed better on FK and HSK. \textbf{(c)} and \textbf{(d)} show the t-SNE plot to visualize the feature distribution of ResNet50 and KGDM respectively when embedding the same data, intuitively showing that prior knowledge penalized feature distribution is more separable.}
    \label{fig: cls_wise}
\end{figure*}

Figure~\ref{fig: cls_wise}a and Figure~\ref{fig: cls_wise}b exhibit the ROC curve of KGDM compared to other CNN-based architecture.
In the range of low false positive rate limitation, KGDM tended to have a higher true positive rate.
Figure~\ref{fig: cls_wise}c and Figure~\ref{fig: cls_wise}d intuitively illustrate how KGDM came to its superiority via a 2D t-SNE scatter plot, where each point represented one sample.
Figure~\ref{fig: cls_wise}c exhibits that the points of different classes intersect with each other with fuzzy boundaries.
As shown in Figure~\ref{fig: cls_wise}d, KGDM groups the features from different classes into ``lines'' with larger margins between different classes as it is constrained by EDL loss that results in the clear boundaries.
We can notice that there exists a small part of the samples intersect with each other at the central areas because KGDM can not find sufficient evidence for a certain class and say \emph{``I do not know''} for these cases.
For a more comprehensive study, we reported the ablation study on the architecture modification, uncertainty threshold, and additional loss in Supplementary Table~\ref{tab:ablation}.

\subsection*{Interpretation of manifestation prototypes and classification procedure}
\label{sec:interpretation}

\begin{figure*}[ht]
    \centering
    \includegraphics[width=\linewidth]{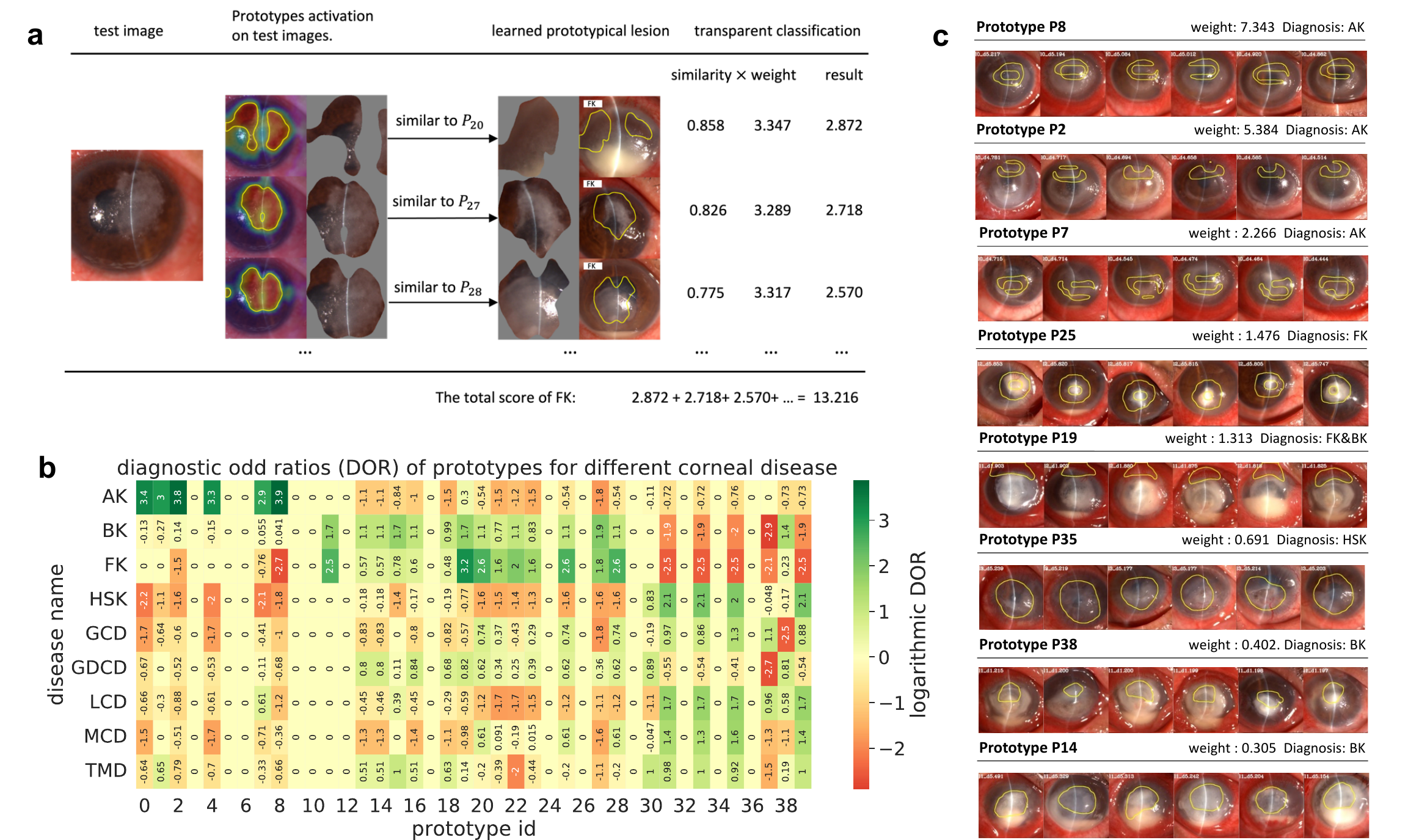}
    \caption{\textbf{Visualized reasoning process and interpretation of learned prototypes.}
    \textbf{a.} A example of the visualized reasoning process showing an HSK patient who was classified through similarities to HSK-related manifestation prototypes. The prototypes are illustrated with retrieved training samples. The high similarity area is circled with yellow curves. 
    \textbf{b.} Quantitative evaluation of prototype. Logarithmic scale of diagnostic odds ratios between 40 prototypes and 9 diseases in SRRSH prospective validation dataset, measuring the correlation between the learned diagnostic pattern and the diseases (values are set to 0 if their 95\% confidence interval spanned over 0).
    \textbf{c.} Qualitative evaluation of learned prototypes. Representative regions of learned prototypes are shown inside the yellow contour, which contained valid and stable signs benefiting the diagnosis of target diseases. The corresponding diagnostic odd ratio (DOR) is calculated on the prospective validation dataset.
    } 
    \label{fig:interpretation}
\end{figure*}

Figure \ref{fig:interpretation}a illustrates the transparent classification process of an input HSK image.
The top 3 prototypes are listed according to their contributions to the target class prediction in descending order.
For each prototype, its activation region is visualized on the test image. Moreover, the most similar diagnostic pattern in the training set is retrieved to show which training sample accounts for this prototype.
In practice, the ophthalmologists could understand the sample in Figure~\ref{fig:interpretation}a as \emph{``KGDM classified the case as FK because these circled regions were looked similar to the training samples that have been diagnosed as FK''}.
More Specifically, the first cropped area in the image has a similarity of 0.858 to the learned diagnostic protoype $P_{20}$, and the pattern $P_{20}$ has a weight of 3.347 in diagnosing FK, so this region contributes to FK with a score of $ 0.858 \times 3.347 = 2.872$. The rest of the proposed patterns contribute $0.422, 0.381, \dots$ correspondingly, so the total prediction score for FK is their sum, 13.216.
We also report the accuracy of retrieved training samples in Supplementary Table~\ref{tab:retrieved_samples_accuracy} and show more examples in Supplementary Figure~\ref{fig:ext_interp}.
The results show that the retrieved samples are highly consistent with the target class, which indicates that the learned prototypes are meaningful and interpretable.

Figure~\ref{fig:interpretation}b shows the logarithmic scale of the evaluated DOR scores of 40 prototypes to 9 kinds of subsequently confirmed ground truth diagnostic results on SRRPV datasets, including 4 targeted subclasses of infectious keratitis and 5 other corneal diseases. The 95\% confidence interval and the statistical significance of DOR scores are shown in Supplementary Table~\ref{tab:dor_scores}.
We set the DOR value to 0 if the 95\% confidence interval spanned over 0, which means the correlation was not statistically significant.
$P_0,P_1,P_2,P_4,P_7,P_8$ have high DOR with AK, where the highest 2 prototypes were $P_2$ with DOR 43.51 (95\% CI 37.7746-50.1081) and $P_8$ with DOR 48.11 (95\% CI 41.7572-55.4349).
For BK, KGDM learned $P_{11},P_{15},P_{19}$, where $P_{19}$ had a DOR of 5.40 (95\% CI 3.9548-7.3619) ,
$P_{22},P_{25},P_{28}$ with FK, where the highest $P_{25}$ had a  DOR of 13.40 (95\% CI 11.6998-15.3558)
and  $P_{31},P_{33},P_{35},P_{39}$ with HSK, where $P_{35}$ had a DOR of 7.33 (95\% CI 5.9756-8.9924).
The results have shown the effectiveness of the learned prototypes when they were applied to real-world diagnosis.

Figure~\ref{fig:interpretation}c displays the prototypes that have high classification weights along with the corresponding prototype index and disease.
Surprisingly, without predefined annotations, the representative regions have shown consistency with the clinical experience of some ophthalmologists.
$P_2, P_7, P_8$ exhibited a high correlation with AK; accordingly, their visual features showed a ring-shaped stromal infiltration, a characteristic clinical sign of AK \cite{Maycock2016,Juarez2018}. Epithelial defects in $P_8$ are also consistent with the clinical experience of AK diagnosis\cite{Maycock2016}.
In $P_{25}$, the corneal lesions exhibit white infiltration with feathery borders, which is an important feature of FK \cite{Ansari2013, Thomas2013}.
In HSK, the corneal infiltration is often accompanied by neovascularization, which has been exactly shown in $P_{35}$.
In $P_{19}$, the severe conjunctival injection can be observed, which are correlated with both BK and FK~\cite{Cabrera-Aguas2022}.

\section*{Human-AI collaborative diagnosis performance}
\begin{figure*}
    \centering
    \includegraphics[width=\linewidth]{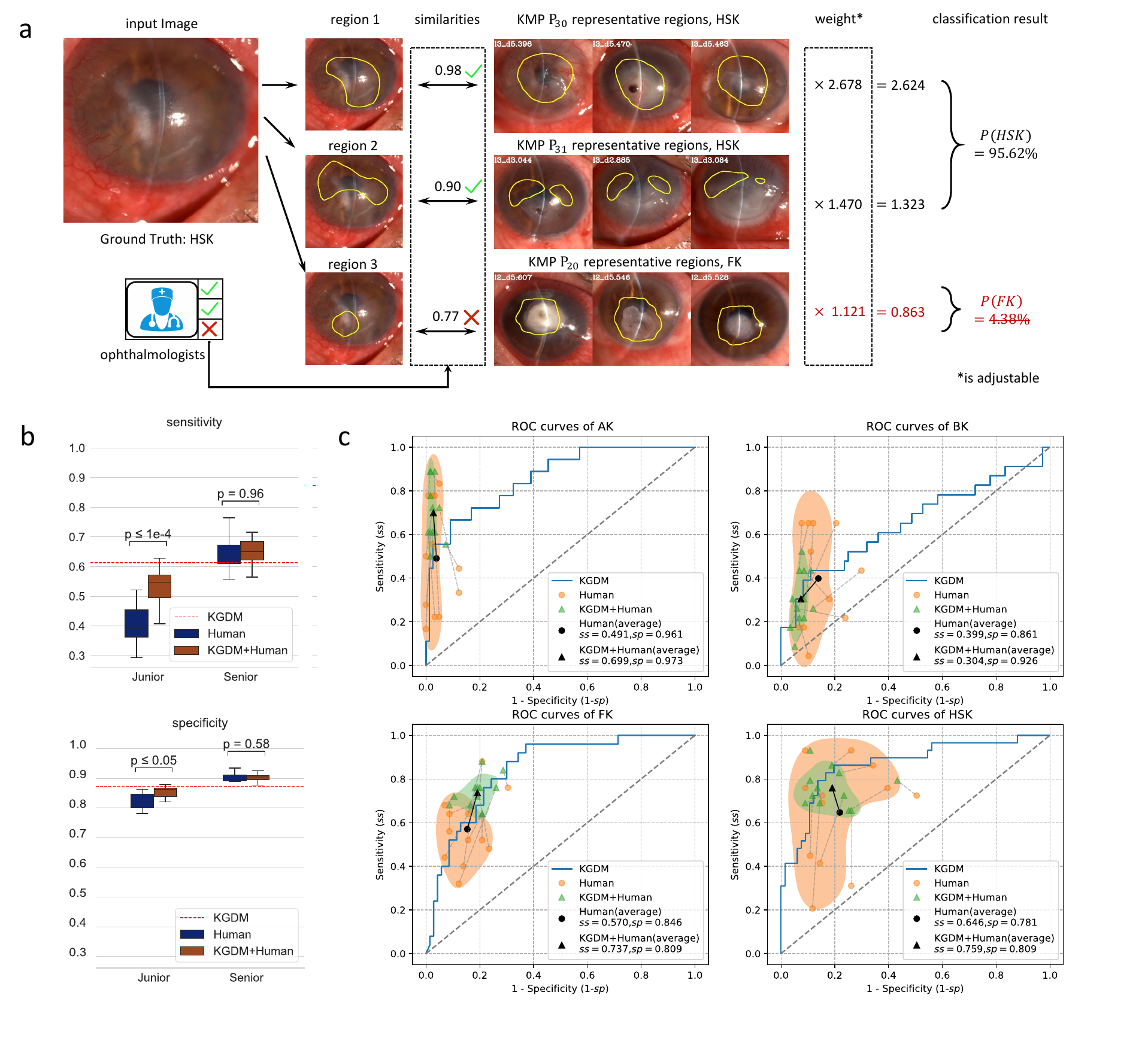}
    \caption{\textbf{The performance of the Human-XAI collaboration test}
    \textbf{(a)} the Human-AI collaboration diagnosing workflow based on sample-level interpretation and interactive interface.
    \textbf{(b)} comparing the diagnosis performance improvement between the two groups, ``Junior'' and  ``Senior'', before and after collaboration with KGDM.
    \textbf{(c)} the ROC curves comparing the effects of each ophthalmologist before and after collaboration with KGDM. The orange dots represent the performance of each tester in the first step (Human), the green triangles represent its performance using KGDM, and the black dots and triangles represent the mean value of the two groups, respectively.} 
    \label{fig:KGDM+human}
\end{figure*}

As shown in Table \ref{fig:KGDM+human}, among the 12 ophthalmologists, the average diagnostic performance of Humans was $ss=52.65\%, sp=86.22\%$.
With the collaboration, the average diagnostic performance of KGMD+Human was $ss=62.47\%,sp=87.91\%$, which was higher than both Humans and KGDM, demonstrating a potential improvement of diagnosis performance in the clinical environment.
Figure \ref{fig:KGDM+human}b) illustrates the subgroup performance.
The results demonstrated a noteworthy promotion of $ss=+17.30\% (P<0.01),sp=+3.14\%(P<0.05)$ for the "Junior" group, while the promotion on the group "Senior" was not significant $ss=+2.34\% (P=0.95),sp=+0.18\%(P=0.58)$.
This result indicates that KGDM can compensate for the less experienced ophthalmologists more, while its benefits for senior ophthalmologists need further investigation.
As shown in Table \ref{tab:KGDM+human}, comparing to the previous average $\kappa^F=0.3067$ of 12 ophthalmologists, collaborating with KGDM achieved $\kappa^F=0.546$, which was $78.02 \%$ higher.
In the two subgroups "Junior" and "Senior", the promotions were still significant with $0.5287$ vs. $0.2064$ and $0.6346$ vs. $0.4861$ respectively.
These results demonstrate that collaborating with KGDM can significantly improve the agreement between ophthalmologists.

\begin{table}[ht]
    \caption{Average ratings of ophthalmologists in classifying infectious keratitis}
        \begin{threeparttable}
            \resizebox{\textwidth}{!}{
                \begin{tabular}{clcccc|c} 
                    \hline
                    \rowcolor{Dark}
                    
                     &  & \multicolumn{2}{c}{Sensitivity}& \multicolumn{2}{c|}{ Specificity } &  \\
                    \rowcolor{Dark}
                    \multirow{-2}{*}{Group}& \multirow{-2}{*}{Method} & mean & std & mean & std & \multirow{-2}{*}{Fleiss Kappa\tnote{1}} \\
                    \hline
                    \multirow{4}{*}{Junior(< 5 year)}& KGDM  &60.57\% &- & 86.73\%&- &-\\
                    & Human & 40.16 \% & 3.32 \% & 82.26 \% & 3.22 \% & 0.2064  \\
                    & KGDM+Human & 57.46 \% & 4.65 \% & 85.40 \% & 2.27 \% & 0.5287  \\
                    & $\Delta$ & +17.46\%  & -  & +3.14 \%  & -  & +0.3223  \\
                    \hline
                    \multirow{4}{*}{Senior(> 5 year)}& KGDM  &60.57\% &- & 86.73\%&- &-\\
                    & Human & 65.13 \% & 6.26 \% & 90.25 \% & 2.53 \% & 0.4861 \\
                    & KGDM+Human & 67.47 \% & 4.81 \% & 90.43 \% & 1.96 \% & 0.6346  \\
                    & $\Delta$ & +6.15\%  & -  & +3.13 \% & -  & +0.1485  \\
                    \hline
                    \multirow{2}{*}{All testers \tnote{2}} & Human & 52.65 \% & 13.89 \% & 86.22 \% & 5.00 \% & 0.3067  \\
                    & KGDM+Human & 62.47 \% & 6.90\% & 87.91\% & 3.31 \% & 0.5460 \\
                    & $\Delta$ & +9.82\%  & -  & +1.69 \% & -  & +0.2393  \\
                    \bottomrule
                \end{tabular}}

            \begin{tablenotes}
                \footnotesize
                \item[1] Fleiss' kappa, to assess the reliability of agreement between a fixed number of raters when classifying items. 
                \item[2] 12 ophthalmologists were recruited for evaluating collaboration with KGDM, separated into two subgroups Junior (n=6) and Senior (n=6).
            \end{tablenotes} 
        \end{threeparttable}
        
        \label{tab:KGDM+human}
    \end{table}

To compare the performance of each ophthalmologist before and after collaboration, we plot the ROC curve in Figure~\ref{fig:KGDM+human}c, where each orange dot represents the performance of each tester in the first step (Human) and the green triangle represents his performance using KGDM  (KGDM+Human). The orange and green points of the same tester are connected with a gray dash line to show its promotion.
The black dots and triangles represent the mean values of the two groups, respectively. On AK, FK, and HSK, the sensitivity ($ss$) has been significantly improved with the mean promotion of +20.83\%, +16.67\%, and +11.21\%, respectively. On BK, the specificity ($sp$) has been significantly improved with the mean promotion of +6.5\%.
The 2D kernel density estimation~\cite{terrell1992variable} is also applied to show the distribution of diagnostic results (orange and green areas represent Human and KGDM+Human). The green areas are significantly smaller than orange ones, which indicates the diagnosis variance has become smaller and KGDM can improve the consistency among ophthalmologists.

Based on Figure~\ref{fig:KGDM+human}c, KGDM+Human exhibited significantly higher $ss$ and $sp$ compared to Humans and KGDM alone in the diagnosis of AK. In terms of BK, KGDM+Human outperformed humans alone in $sp$ for 11 out of 12 testers, and $ss$ for 7 out of 12 testers. Moreover, for diagnosing FK, KGDM+Human exhibited higher $ss$ than Humans alone for all testers, and higher $sp$ for 6 out of 12 testers. Finally, for HSK, KGDM+Human demonstrated higher $ss$ for 6 out of 12 testers, higher $sp$ for 4 out of 12 testers, and no significant change in performance for 2 out of 12 testers. These results indicate that collaborating with KGDM can help ophthalmologists improve their diagnostic ability in diagnosing infectious keratitis in the clinical environment.

%% file: _discussion.tex
\section*{Discussion}
It is aspirational to achieve complementary performance that surpasses both humans and AI in real-world applications~\cite{Bansal2021, Inkpen2022a}. However, previous attempts in medical image analysis are limited to collectively fusing the decision of AI and clinicians \cite{Patel2019} or taking over the decision~\cite{Ahn2022, Leibig2022}, where neither side can actively engage in the decision-making process of the other because of the \emph{``black box''} nature of deep learning~\cite{Lai2021}.
Self-explainable models like ProtoPNet~\cite{Chen2019} allow users to understand the model's reasoning process.
Though there are some existing works that migrate ProtoPNet to medical image diagnosis~\cite{Kim2021, Barnett2021}, their performance is limited compared to the \emph{``black box''} counterparts and the interpretation have not been validated in clinical environment.
Recently, the astonishing success of GPT4 \cite{OpenAI2023} and Segment Anything \cite{Kirillov2023} suggest a promising solution by incorporating human prompts into the model reasoning through human-AI interaction.
This is coherent with what our proposed KGDM does in the collaborative diagnosis. By integrating the diagnosis process and expert experience, our self-explainable model offers clinicians a transparent classification process and provides an interactive interface that enables them to intervene in the reasoning process.
The test result of collaborative diagnosis in Table~\ref{tab:KGDM+human} and Figure~\ref{fig:KGDM+human} indicate the significant improvement of junior ophthalmologists, along with an enhanced consistency among the diagnostic opinions of ophthalmologists. The results suggest that complementary performance between clinicians and AI in decision-making based on medical image analysis is achievable via self-explainable models.

Our study is distinguished from existing researches that diagnosing infectious keratitis with AI~\cite{Ghosh2022, Koyama2021,Kuo2020,Kuo2022,Li2021a,Redd2022, Redd2022a}. Most of them directly applied general network architecture, \emph{e.g.} VGG~\cite{Simonyan2015}, ResNet~\cite{He2016} and DenseNet~\cite{Huang2017a}, which implicitly map input image to binary classification results for determining wether it belongs to a certain disease. In contrast, our KGDM explicitly learns and visualizes the diagnostic patterns from the data and provides the interpretation of the reasoning process as shown in Figure~\ref{fig:interpretation}a and Supplementary Figure~\ref{fig:ext_interp}. The visualized prototypes in Figure~\ref{fig:interpretation}c can be interpreted with the expert knowledge about IK and can be used to identify different subtypes, \emph{e.g.} the typical \emph{``ring infiltration''} of AK and \emph{``feathery borders''} of FK. They are automatically learned from training data and their activation are also validated in the prospective test. The high odd ratios in Figure~\ref{fig:interpretation}b and Supplementary Table~\ref{tab:dor_scores} indicate the learned prototypes are effective image biomarkers that provide auxiliary information for clinicians. What is more, ophthalmologists were recruited to compare their performance against AI algorithms in previous work~\cite{Xu2021, Redd2022a}, which suggests the existing \emph{``black box''} systems aim to replace a doctor~\cite{Barnett2021}. In contrast, we performed the comparison on a simulated clinical environment and additionally evaluated their performance as ``collaborated teammates'', suggesting that KGDM is a decision-aid rather than a decision-maker. The collaborated results demonstrates collaboration with self-explainable AI is a promising new diagnosis paradigm for infectious keratitis. In contrast to the previous paradigm, where junior ophthalmologists had to seek opinions from more experienced colleagues to reduce false decision. The new paradigm equips ophthalmologists with the reasoning process of AI, thereby facilitating the diagnostic process by providing informative AI-based diagnostic biomarkers. This paradigm might be extended to other diseases where experienced doctors are limited, opening up possibilities for enhanced diagnostic capabilities from data-driven intelligence.

Although the performance analysis in Table~\ref{tab:performance} and Figure~\ref{fig: cls_wise} indicate that KGDM surpasses the performance of previous methods, the trade-off between interpretability and classification performance~\cite{Rudin2019} still exists in KGDM. From the perspective of feature extraction, we mainly evaluate a representative architecture \emph{ResNet50} for interpretation. Although it is conceivable that replacing it with better backbones like ViTs~\cite{Liu2021a} or adding more layers like \emph{ResNet152} could potentially yield better performance,the increased feature complexity and excessively large receptive field might harm interpretability. The ablation experiments in the Supplementary Table\ref{tab:ablation} show that the improvement in classification performance primarily from evidential loss~\cite{Sensoy2018}, while the addition of a prototype layer weakens the performance. Systematically searching all potential encoder architecture and hyperparameters for the optimal combination remains to be future work. From the perspective of training data, to ensure high-quality reference samples are provided to users, we conducted strict quality control during the data collection phase. While this potentially introduces selection bias and might harm generalizability, in the diagnosis of IK, the excluded samples lack features understandable by humans. Including them in the training set might lead to learning spurious correlated prototypes, undermining users' trust to the model. What is more, the 12 ophthalmologists who participated in the collaboration test were recruited from two medical centers. To get more rigorous conclusions, cross-regional research should be conducted with large-scale trials in future work. Although our proposed method has established a bridge to some extent between image-processing AI and medical professionals, the current collaboration is merely a preliminary exploration of the feasibility of human-AI complementarity. Building upon this foundation, the considerable potential of human feedback loops could be further investment and exploration~\cite{zotero-1674}.




In conclusion, we have presented a self-explainable model guided by expert knowledge, which demonstrated superior performance through human-AI complementarity. Our research is in line with the perspective of \emph{``Learning From Experts, Examples, and Experience''}~\cite{Chen2022}, and provided a feasible basis for promoting performance from human feedback through interpretability and interactivity, by initiating the communication between image classification AI and clinicians.

%% file: _methods.tex
\newpage

\section*{Methods}

\subsection*{Datasets}

In this research, a total of 5 datasets were used for model training, validation, testing, and the Human-XAI collaboration test, including the SRRSH training dataset (SRRT), the SRRSH prospective validation dataset (SRRPV), the XS hospital testing dataset (XS), the online available publication testing dataset (PUB), and the Human-XAI collaboration test dataset (HXACT).

\textbf{SRRSH training dataset} The SRRSH training dataset for this study is a subset of the dataset, which includes 129853 clinical digital images of 89 categories of corneal diseases. These clinical images were taken by slit lamp microscopy during the time period of May 2004 to 2021 in the Department of Ophthalmology, Sir Run Run Shaw Hospital, School of Medicine, Zhejiang University. Two types of equipment have been used to take the images, that is, Zeiss slit lamp microscope SL 130 (Carl Zeiss Meditec AG, Germany), integrated with the SL Cam for imaging module, providing each image with a resolution of $1024 \times 768$ pixels; and Topcon slit lamp microscope (TOPCON Corporation, Japan), affiliated with digital camera Unit DC-1 offering an image resolution of $1740 \times 1536$ pixels or $2048 \times 1536$ pixels. 

For all the cases in the raw dataset, we restricted the visit time to between January 2014 and January 2021. Cases with diagnoses of other than the four categories of infectious keratitis (AK, BK, FK, and HSK) were excluded. We reviewed the medical records of each case, including slit lamp microscope images, examination reports, and medical documents that record clinical manifestations and treatments. Cases with complete medical records as described above were included.
All the diagnoses were reconfirmed by ophthalmologists of the Department of Ophthalmology, Sir Run Run Shaw Hospital, School of Medicine, Zhejiang University. The diagnosis of infectious keratitis was corroborated by at least two pieces of the following evidence: (1) the responsiveness to appropriate drug treatment, (2) the detection of specific pathogens, (3) typical clinical characteristics. The responsiveness to drug treatment was confirmed after standard treatment. Antibiotics, anti-fungal drugs, and anti-viral drugs, respectively, were used to treat bacterial, fungal, and HSV keratitis. Chlorhexidine, with or without anti-fungal drugs, was used to treat acanthamoeba infections. Smear examination and culture of corneal samples collected from patients suspected of having BK, FK, or AK were made to detect the specific pathogens. PCR was used to diagnose HSV infection. In addition to the treatment response and laboratory tests for pathogenic microorganisms, typical clinical characteristics of the corneal infection were reviewed from the slit lamp microscope images (as shown in Figure.\ref{fig:exampleIK}). Cases with an unclear diagnosis or co-infection, where more than one pathogen was involved in a single case, were excluded. After the multi-step filtrating procedure (Extended Data Figure.\ref{fig:inclusion and exclusion}), a total of 671 cases with a definite diagnosis of either one of the four categories of infectious keratitis were included in our study, among which were 336 HSK cases, 109 BK cases, 178 FK cases, and 48 AK cases. We performed an additional review of the images to exclude those who represent the non-acute stage and lost typical visual features of relevant infectious keratitis. Images with blue light illumination to view fluorescein-stained corneas were also excluded.

In total, 3223 qualified images from 591 patients were included in the SRRSH training dataset (333 images of acanthamoeba keratitis, 527 images of bacterial keratitis, 977 images of fungal keratitis and 1386 images of herpes simplex virus keratitis). 

\textbf{SRRSH prospective validation dataset }
For the SRRSH prospective validation dataset, we prospectively enrolled 336 patients who visited the Department of Ophthalmology, Sir Run Run Shaw Hospital, School of Medicine, Zhejiang University for "blurred vision with eye irritation or eye pain" between May 2021 and June 2022. Additional medical information affiliated with each image reviewed, the ground truth labels were annotated by the same ophthalmologists as the SRRSH training dataset. The SRRSH prospective validation dataset included 88 images of AK, 92 images of BK, 138 images of FK, 153 images of HSK, 554 images of other corneal diseases (e.g., corneal dystrophy, recurrent corneal erosion, Terrien marginal degeneration), and 146 images of corneas without significant abnormalities.

\textbf{XS hospital testing dataset} 
The XS hospital testing dataset is a subset of the dataset which includes clinical images taken by slit lamp microscopy during the time period of March 2005 to 2021 in the Department of Ophthalmology, Xiasha Branch of Sir Run Run Shaw Hospital, School of Medicine, Zhejiang University. After being selected and annotated by the same criteria as the SRRSH training dataset, a total of 138 images were included in this dataset, consisting of 26 images of AK, 31 images of BK, 31 images of FK, and 50 images of HSK.

\textbf{Online available publication testing dataset}
We also collected 66 images for the PUB dataset, including 15 images of AK, 11 images of BK, 20 images of  FK, and 20 images of HSK. We searched Google for slit lamp microscopic images using the keywords "keratitis", "amoeba keratitis", "bacterial keratitis", "fungal keratitis" and "herpes simplex virus stromal keratitis". Images from other articles or from official hospital websites were preferred in our dataset because of their greater reliability. All the images and their labels were rechecked by the ophthalmologists.

\textbf{ Human-XAI collaboration test dataset}
The  Human-XAI collaboration test dataset was collected in the same way as the SRRSH prospective validation dataset; it contained 18 images of AK, 23 images of BK, 25 images of FK, 29 images of HSK, 32 images of other corneal diseases, 13 images of corneas without significant abnormalities.

\subsection*{Knowledge-guided Data-driven Model}
\textbf{Architecture and notions.} The network architecture of KGDM inherits from ResNet50 \cite{He2016} with an additional branch for the transparent classification which introduces the design of \textit{prototype layer} \cite{Chen2019} as shown in Figure \ref{fig:KGDM}b.
The overall model consists of three parts, including a feature map extractor $F(\cdot)$, a distribution predictor $g \circ l_{max}$, and the prototype layer $\mathbf w \circ l_{max} \circ \Phi_{\mathbf P}\left(\cdot\right)$,
where $g \circ l_{max}$ means a fully connected layer $g$ following a max pooling layer $l_{max}$.
$\Phi_{\mathbf P}\left(\cdot\right)$ outputs the similarity between the learnable prototype module $\mathbf P$ and the input feature map.
$\mathbf{w}$ is a learnable classification weight matrix.
An input image $X$ is firstly embedded into a feature map $F(X) \in \mathbb R^{h \times w \times d}$, where $h$,$w$ and $d$ are the height, width, and feature channels of $F(X)$ correspondingly. Given $l_{max}:\mathbb R^{h \times w \times d} \rightarrow \mathbb R^d$ and $g:\mathbb R^{d} \rightarrow \mathbb R^k_+$, the distribution predictor predicts an $\mathbf \alpha=(\alpha_1,\dots,\alpha_k)$ for each category as follows,
\begin{equation}
    \mathbf \alpha = g\circ l_{max} \circ F(X) + 1,
\end{equation}
where $k$ is the number of categories.
In the transparent classification branch, given $\mathbf P \in \mathbb R^{k \times m \times d}$, $\Phi_{\mathbf P}: \mathbb R^{h \times w \times d} \rightarrow \mathbb R^{h \times w \times k \times m}_+$, and $\mathbf{w} \in \mathbb R^{k \times m}_+$, the final predicted classification scores are denoted as $\mathbf \tau \in \mathbb R^k_+$, where the score $\mathbf \tau_i$ for the category $i$ is calculated as:
\begin{equation}
    \mathbf \tau_i = \sum_{j=1}^{m} \mathbf w_{ij} \left[l_{max} \circ \Phi_{\mathbf P}\left(F(X)\right)\right]_{ij},
\end{equation}
where $m$ is the number of the prototypes for each category.

The expert knowledge is integrated into the above architecture in three phases:
First, we mimic the expert diagnosis process, which summarizes diagnostic patterns from the historical images and utilizes them as experiential knowledge in clinical diagnosis. To achieve transparent classification, the classification is calculated by the weighted sum of similarity. The prototype module $\mathbf P$ that memorizes the feature vectors of prototypes is updated iteratively. We have designed a visualization strategy and retrieved the most similar training images to interpret the similarity, called \textit{interpretable modeling}.
Second, the feature map extractor is regularized by regarding the predicted $\mathbf \alpha$ as an agent during training, called \textit{modeling uncertainty based on expert prior experience.}
Third, after training, KGDM provides an interactive interface to directly incorporate expert knowledge into the final classification, called \textit{Human-AI collaborative diagnosing.}
We will introduce the above three strategies as follows:

\textbf{Interpretable modeling by referring to the expert diagnosing process.} \label{sec:phase2}
Since KMPs are summarized from the historical cases of ophthalmologists and used to assess their similarity to diagnose IK for new patients, a prototype module is introduced to mimic this diagnosis process. The module memorizes the prototypes of keratitis as embedding vectors and performs the classification with the weighted sum of cosine similarity between these vectors and the feature maps of the input images. The classification errors are minimized by iteratively updating the prototype vectors and the corresponding network weights. Assuming that the feature map might embody possible evidence at each position, a visualization policy for direct comprehension of the diagnostic pattern is designed as shown in Figure \ref{fig:KGDM}c, see Eq \ref{eq:inferring}. Based on the visualization, we can perform both sample-level and class-level interpretations. For sample-level interpretation, a cosine similarity map is calculated between the feature map and each learned prototype vector. Scaled to the same size as the input image, it becomes a heatmap to show which parts of the image are similar to the learned prototype. The high similarity area is circled within a yellow contour. The maximum similarities of all prototypes are summed with classification weights for target diseases.
The sample-level interpretation provides a way to understand which exact regions of the image are being used for classification and how much they contribute to the final classification.
For class-level interpretation, after training, each prototype vector would traverse the whole training dataset and identify the closest image regions as its representative patterns
The representative patterns have maximum similarities to the prototype vector among all training samples.
We visualized the closest regions of each prototype for the 4 target diseases and printed the corresponding classification weights.
The class-level interpretation provides a way that ophthalmologists could understand what kind of patterns the AI has learned for diagnosing certain diseases and how important they are for diagnosing the diseases.

A transparent classification branch is designed to mimic the inference process of experienced ophthalmologists.
The experienced ophthalmologists usually summarize the diagnostic patterns, namely KMP, from a large number of patients they have seen before, and compare them to a new patient for diagnosis.
To mimic this process, we first design a module $\mathbf P \in \mathbb R^{k \times m \times d}$ for $k$ subclasses of IK and let each of them have $m$ prototypes.
As shown in Figure \ref{fig:KGDM}c, the input feature map $F(X)$ is dissected according to position $p,q$, and \textit{cosine} similarities are calculated for the prototype feature vector $\mathbf P$ one by one.
This process is denoted as $\Phi_{\mathbf P}$, which outputs a tensor $\mathbf S \in \left[-1,1\right]^{k \times m \times h \times w}$ that contains all similarities:
\begin{equation}
    \Phi_{\mathbf P}(F(X)): \mathbf S_{ijpq}=\frac{\mathbf P_{ij} F(X)_{pq}}{\| \mathbf P_{ij}\Vert \| F(X)_{pq}\Vert},\text{where } 1 \leq i \leq k, 1 \leq j \leq m, 1 \leq p \leq h, 1 \leq q \leq w.
    \label{eq:inferring}
\end{equation}
Once $\mathbf S$ is obtained, we can obtain a similarity map $\mathbf S^{X}_{ij} \in \left[0,1\right]^{h \times w}$ for the $j$-th prototype of the $i$-th subclass, in which the greatest similarity is $s^X_{ij}$.
By interpolating and scaling $\mathbf S_{ij}$ to the same size as input $X$ and drawing a yellow contour along with the similarity threshold equal to 0.5 as shown in Figure \ref{fig:KGDM}c, we can visually interpret $s^X_{ij}$ as: for input image $X$, the learned diagnostic pattern represented by prototype $\mathbf P_{ij}$ appears in the regions inside the contour,
where the maximum similarity is $s^X_{ij}$.
The classification score $\mathbf \tau_i$ for class $i$ is the weighted sum of $s^X_{ij}$ that is directly calculated as:
\begin{equation}
    \mathbf \tau_i = \sum_{j=1}^{m} \mathbf{\mathbf w}_{ij} s^{X}_{ij}.
    \label{eq:trans_cls}
\end{equation}
Applying the $softmax$ operation on the $ \mathbf \tau_i$, we obtain the predicted probability of disease $i$ for input images $X$:
\begin{equation}
 p(i \mid X)=e^{ \mathbf \tau_i} / \sum_{j=1}^{k} e^{ \mathbf \tau_j} 
\end{equation}
With this process, KGDM can diagnose the disease based on the cosine similarity between the learned prototypes and the features of the input image.
And the reasoning process is transparent due to the visualization of prototype and the weighted classification.

The loss function for classification is defined as follows:
\begin{gather}
    \mathcal{L}_{trans}=-\frac{1}{n} \sum_{X \in D} \sum_{i=1}^{k} \mathbf y_{i}^{X} log(p(i \mid X)) +\lambda_1 \mathcal{L}_{clst} + \lambda_2 \mathcal{L}_{sep}\\
    \mathcal{L}_{clst}=-\frac{1}{n} \sum_{X\in D}  \sum_{i=1}^{k} \mathbf y_{i}^{X} \min_{j=1}^{m} s_{ij}^{X} \\
    \mathcal{L}_{sep}=\frac{1}{n} \sum_{X\in D}  \sum_{i=1}^{k}(1- \mathbf y_{i}^{X}) \max_{j=1}^{m} s_{ij}^{X}
\end{gather}
where $D$ denotes the training dataset that have $n$ samples. The term $\mathcal{L}_{clst}$ maximizes the minimum $s^X_{ij}$ of the prototype of the ground truth class $i$ and $ \mathcal{L}_{sep}$ minimizes the maximum $s^X_{ij}$ of the prototypes that is not the ground truth $i$.

\textbf{Modeling uncertainty based on expert experience.} Considering the existing uncertainty in making a diagnosis opinion, we introduced an auxiliary uncertainty estimation task in KGDM. In particular, a Dirichlet distribution (referring to Eq. \ref{eq:dirichlet}) is placed on class probabilities to model the prediction variances. To predict the distribution, another branch is incorporated to take the embedded feature maps as input, which is fed into a fully connected layer after max pooling. The branch outputs non-negative parameters $\mathbf{\alpha}$ of Dirichlet distribution as shown in Figure \ref{fig:KGDM}b.
In this way, the likelihood of each possible categorical distribution over the labels of the classified samples is represented by the Dirichlet distribution.
According to expert experience, well-learned features should be common in the same subclass but distinguishing among different subclasses.
We designed a loss consisting of the prediction error and the variance of the predictive distribution, whose optimization benefits reducing the influence of confounding patterns.
The expected prediction probability is calculated based on the estimated $\mathbf{\alpha}$ on all categories, as shown in Eq. \ref{eq:prob_c}.
For the prediction, the higher the $\mathbf{\alpha}$, the more confident the model is. Specifically, when $\mathbf{\alpha}$ is an all-one vector, it suggests the model is totally uncertain as the predicted probability of each subclass of IK is equal. In addition, an `ideal' distribution corresponding to the uniform distribution is employed to construct an auxiliary Kullback-Leibler divergence between it and the predicted Dirichlet distribution, which is added to the overall loss function to eliminate the adverse inter-class correlation effect brought by the totally uncertain samples, as shown in Eq. \ref{eq:kl}. This turns the prior experience into a constraint that makes the learned feature space more stable and reliable.

We predict the classification events distribution with a Dirichlet distribution parameterized by $\mathbf \alpha$ rather than directly predicting the probability for each category.
After learning the predicted $\mathbf \alpha$ for a $k$-classification, the probability density function of every possible probability assignment $\mathbf p=(p_1,\dots,p_k)$ is:
\begin{equation}
    D_{\mathbf \alpha}(\mathbf p)=\frac{\varGamma(\alpha_0)}{\sum_{i=1}^k\varGamma(\alpha_i)} \prod_{i=1}^{k}p_i^{\alpha_i-1}, \text{where } \mathbf p \in \mathcal S.
    \label{eq:dirichlet}
\end{equation}
Here $\mathcal S$ is the $k-1$ dimensional simplex, where
$\forall \mathbf p \in \mathcal S, \| \mathbf p \Vert_1 =1,p_i\in [0,1]$.
According to the characteristics of Dirichlet distribution, the expected probability of class $i$ is:
\begin{equation}
    \mathbb E(p_i)=\frac{\alpha_i}{\alpha_0}, \alpha_0 =\sum_{i=1}^k\alpha_i
    \label{eq:prob_c}
\end{equation}
This learning process is to minimize the classification error between the predicted expected probability and the ground truth given by one-hot training label $\mathbf y \in \{0,1\}^k$.
The loss function is written as follows:

\begin{equation}
    \begin{split} \mathcal{L}^{err}(\mathbf y, \mathbf\alpha) 
    &= \int_{\mathbf p \in \mathcal S } \| \mathbf y-\mathbf p\Vert_2^2 D_{\mathbf \alpha}(\mathbf p)\,d\mathbf p \\
    &= \mathbb E(\|\mathbf y-\mathbf p\Vert_2^2) \\&= \mathbb E(\sum_{j=1}^{k}(y_{j}- p_j)^2) \\&= \sum_{j=1}^{k}[\mathbb E(y_{j}^2)+\mathbb E(p_j^2)-2\mathbf y_{j} \mathbb E(p_j)] \\
    &=\sum_{j=1}^{k}[\mathbf y_{j}^2+\mathbb E(p_j)^2+Var(p_j)-2\mathbf y_{j} \mathbb E(p_j)] \\
    &=  \sum_{j=1}^{k}[(\mathbf y_{j}-\mathbb E(p_j))^2+Var(p_j)]  \\
    &= \sum_{j=1}^{k}[(\mathbf y_{j}-\frac{\alpha_j}{\alpha_0})^2+\frac{\alpha_j(\alpha_0-\alpha_j)}{\alpha_0^2(\alpha_0+1)}],
    \end{split}
    \label{eq:loss}
\end{equation}

by minimizing $\mathcal{L}^{err}(\mathbf y, \mathbf\alpha) $, the model can learn some diagnostic patterns in a data-driven manner, which can provide an optimal classification performance given a training dataset.
According to Subjective Logic \cite{Josang2016}, the relationship between uncertainty $u$ and evidence is bridged by the following:
\begin{equation}
    \sum_{i=1}^{k}\frac{\alpha_i-1}{\alpha_0}+u = 1, \text{where } \alpha_0 = \sum_{i=1}^{k} \alpha_i,
\end{equation}
For example, in our IK classification problem, given an image of FK, the model may predict $\hat{\mathbf \alpha}=(3.5,4.5,5.5,2.5)$ for AK, BK, FK, and HSK correspondingly, the predicted expected probability of FK is calculated as $ \hat{\mathbb E \left(p_{FK}\right)}=\frac{5.5}{3.5+4.5+5.5+2.5}=0.34375$, and the uncertainty is $\hat u = 1-\frac{12}{16}=0.25$. Extremely, if the model has found `infinite' confidence of FK, then  $\alpha_{FK} \rightarrow \infty $, makes $\sum_{i=1}^{k}\frac{\alpha_i -1}{\alpha_0} \rightarrow 1 $ and $ \hat{u} \rightarrow 0, \hat {\mathbb E \left(p_{FK}\right)} \rightarrow 1$, which indicates that the model is confident for the prediction. In contrast, if none of the diagnostic features is observed, which corresponds to the situation where the distribution predictor $g$ outputs an all-one vector, where $\sum_{i=1}^{k}\frac{\alpha_i -1}{\alpha_0} = 0 $  and $\hat{u} = 1$, indicating that the model is pretty uncertain for its prediction. In this case, the probability of all possible probability assignment $\mathbf p \in \mathcal S$ is uniform where $\hat{\mathbb E \left(p_{AK}\right)}=\hat{\mathbb E \left(p_{BK}\right)}=\hat{\mathbb E \left(p_{FK}\right)}=\hat{\mathbb E \left(p_{HSK}\right)}=\frac{1}{4}=0.25$.

We hope the learned feature is less confounding. In other words, the features are expected to satisfy the condition of the lowest inter-class correlation (inter-class similarities are as low as possible given the same confidence). Next, we prove the condition under which the parameters satisfy the lowest interclass correlation in the Dirichlet distribution.

Given a Dirichlet distribution with a covariance function of $Cov(\alpha_{i}, \alpha_{j}) = \frac{-\tilde{\alpha_{i}}\tilde{\alpha_{j}}}{\alpha_{0}+1} $ where $ i\neq j$ , $\tilde{\alpha_{i}} 
= \frac{\alpha_{i}}{\alpha_{0}}$ and we let $\alpha_{0}$ be a constant.
then
\begin{equation}
    \begin{aligned}
arg min \sum_{i=1}^{k} \sum_{j=1,j\neq i}^{k} |Cov(\alpha_{i}, \alpha_{j})| 
& = arg min \sum_{i=1}^{k} \sum_{j=1,j\neq i}^{k} |-\frac{\tilde{\alpha _{i}}\tilde{\alpha _{j}}}{\alpha_{0}+1} |  \\
& = arg min \sum_{i=1}^{k} \sum_{j=1,j\neq i}^{k} |-\frac{\alpha _{i}\alpha _{j}}{(\alpha_{0}+1)\alpha_{0}^{2}} | 
    \end{aligned}
\end{equation}
Since $\alpha_{i}, \alpha_{j}, \alpha_{0}$ are positive numbers.
\begin{equation}
    \begin{aligned}
arg min \sum_{i=1}^{k} \sum_{j=1,j\neq i}^{k} |-\frac{\alpha _{i}\alpha _{j}}{(\alpha_{0}+1)\alpha_{0}^{2}} | 
& = arg min \frac{1}{(\alpha_{0}+1)\alpha_{0}^2} \sum_{i=1}^{k} \sum_{j=1,j\neq i}^{k} {\alpha _{i}\alpha _{j}} \\
& = arg min \frac{1}{(\alpha_{0}+1)\alpha_{0}^2} \frac{(\sum_{i=1}^{k}\alpha_{i} )^2 - \sum_{i=1}^{k}\alpha _{i}^2}{2} \\
& = arg min \frac{1}{(\alpha_{0}+1)\alpha_{0}^2} \frac{\alpha _{0}^2 - \sum_{i=1}^{k}\alpha _{i}^2}{2} 
    \end{aligned}
\end{equation}
Since the constant is irrelevant to finding the minimum of the expression, we remove the constant, and the problem is transformed into finding $ argmax\sum_{i=0}^{k}\alpha_{i}^2 $. From the Cauchy-Schwarz inequality, we know that $\sum_{i=0}^{n}\alpha_{i}^2 \ge \frac{(\sum_{i=0}^{n}\alpha _{i})^2}{2} $. And since we know that all $\alpha_{i}$ is positive. So, to maximize the function we should let one of the $\alpha_{i}$ values be the maximum value and the rest be the minimum value.

For example, in the previous case of FK, we hope the prediction is $\hat{\mathbf \alpha_{ideal}}=(1,1,4.5,1)$,
which can be explicitly modeled by constructing an `ideal' distribution parameterized by $\mathbf \beta=(1,1,5.5,1)$. An auxiliary loss is added to minimize the Kullback-Leibler divergence between the predicted Dirichlet distribution $D_{\mathbf{\alpha}}\left(\mathbf p \right)$ and the `ideal' distribution $D_{\mathbf{\beta}}\left(\mathbf p \right)$. This turns the prior knowledge into a constraint that makes the feature space more coherent with the expert's prior knowledge.
The auxiliary loss is shown as:

\begin{equation}
    \begin{aligned}
    K L(D_{\mathbf{\alpha}}\left(\mathbf p \right) \| D_{\mathbf{\beta}}\left(\mathbf p \right))=&\int_{\mathbf p \in \mathcal S} D_{\mathbf{\alpha}}\left(\mathbf p \right)(log(D_{\mathbf{\alpha}}\left(\mathbf p \right)-log(D_{\mathbf{\beta}}\left(\mathbf p \right))) \,d{\mathbf p} \\ 
    = & \left\langle\log \Gamma\left(\alpha_{0}\right)-\sum_{k=1}^{K} \log \Gamma\left(\alpha_{k}\right)+\sum\left(\alpha_{k}-1\right) \log p_{k}\right. \\
    & \left.-\log \Gamma\left(\beta_{0}\right)+\sum_{k=1}^{K} \log \Gamma\left(\beta_{k}\right)-\sum\left(\beta_{k}-1\right) \log p_{k}\right\rangle_{D_{\mathbf{\alpha}}\left(\mathbf p \right)}\\
    = & \log \Gamma\left(\alpha_{0}\right)-\sum_{k=1}^{K} \log \Gamma\left(\alpha_{k}\right)-\log \Gamma\left(\beta_{0}\right) \\
    & +\sum_{k=1}^{K} \log \Gamma\left(\beta_{k}\right)+\sum_{k=1}^{K}\left(\alpha_{k}-\beta_{k}\right)\left\langle\log p_{k}\right\rangle_{D_{\mathbf{\alpha}}\left(\mathbf p \right)}\\
    =& \log \Gamma\left(\alpha_{0}\right)-\sum_{k=1}^{K} \log \Gamma\left(\alpha_{k}\right)-\log \Gamma\left(\beta_{0}\right)+\sum_{k=1}^{K} \log \Gamma\left(\beta_{k}\right)\\
    &+\sum_{k=1}^{K}\left(\alpha_{k}-\beta_{k}\right)\left(\psi\left(\alpha_{k}\right)-\psi\left(\alpha_{0}\right)\right),
    \end{aligned}
    \label{eq:kl}
\end{equation}

Note that we intend to penalize $\alpha_i \rightarrow 1,(i\neq k)$, it equals to penalizing the agent distribution $D_{\hat{\mathbf \alpha}}(\mathbf p)$ to uniform distribution $D_{\{1\}^k}(\mathbf p)$, where $\hat{\mathbf \alpha}=(1-\mathbf y_i) \odot (\mathbf \alpha-1)+1$. According to Eq. \ref{eq:kl}, we have got the expert knowledge constraint as below:
\begin{equation}
    \mathcal{L}_{i}^{kl} = KL(Dir(\hat{\mathbf \alpha})) \| Dir(\{1\}^k)=\log \frac{\Gamma\left(\hat \alpha_{0}\right)}{\Gamma\left(k\right)\prod_{i=1}^{k}\Gamma(\hat \alpha_k)} +\sum_{k=1}^{K}\left(\hat \alpha_{k}-1\right)\left(\psi\left(\hat \alpha_{k}\right)-\psi\left(\hat \alpha_{0}\right)\right)
\end{equation}

The training objective is the combination of $\mathcal{L}_{i}^{err}$ and $ \mathcal{L}_{i}^{kl} $ for each sample $i$
\begin{equation}
    \mathcal{L}_{i}^{prior} = \mathcal{L}_{i}^{err} + \lambda_{kl} \mathcal{L}_{i}^{kl},
\end{equation}
where $\lambda_{kl}$ is a gradually increased coefficient from 0 to 1 during training.
All trainable parameters in KGDM are trained jointly with the following loss function:
\begin{equation}
    \mathcal{L}_{total}=\mathcal{L}_{trans}+\mathcal{L}_{prior}
\end{equation}

\textbf{Human-AI complementary diagnosing.} Building upon the designed prototype module, the diagnostic patterns of infectious keratitis can be summarized with the learned embedding vector and the related set of feature map regions, which also correspond to the particular image regions. 
For a test image, each prototype vector can compute a similarity map, where the maximum similarity value is considered as its activation. Since the final prediction is obtained by multiplying the activations of all prototype vectors with the corresponding classification weights, it can be self-explained by visualizing the activations of all the prototype vectors and their contribution.
This not only makes the learned features visually comprehensible but also provides a bridge for the interaction between AI and ophthalmologists. Further, a human-AI complementary diagnosing interface is well-designed to incorporate expert opinion into the diagnosing process. In particular, the ophthalmologist may intervene in the reasoning process by editing some prototypes out to neglect their corresponding contributions. On the other hand, the manifestation prototypes of keratitis can be regarded as a kind of AI-based predictive factor, namely imaging biomarkers in some literature \cite{Zeydan2020}, for diagnosing specific infectious keratitis.

In this section, we will introduce how physicians incorporate their prior knowledge into KGDM based on the class and sample-level interpretation, including \textit{global adjustment} and  \textit{local discarding.} After training, we obtain 10 prototypes for 4 diseases, which are named from $P_0$ to $P_{39}$. For interpretation, we retrieve the most similar training images from different patients as the representative images $\mathbb{I}_{\mathbf P_{ij}} $ for each prototype vector $\mathbf P_{ij}$. To figure out the critical regions of these images, we use bilinear interpolation to scale the corresponding similarity map $S^{P_i}$ to the same size as the input images and then draw a contour line (in yellow color) along pixels whose cosine similarity is 0.5. The regions inside the contour represent critical visual information for diagnosis whose embedding representation vector is highly close to the prototype vector, and naturally, their visual perception by KGDM is also highly coherent. 

As the prototypes is automatically learned from retrospective training data, the potential over-fitting may result in a spurious correlation.
Considering the activation of prototypes as a new kind of diagnostic marker,
the diagnostic odd ratio (DOR) is employed to evaluate the correlations between diseases and prototypes on the prospective validation dataset.
For a prototype $P_i$ and a disease $c$, its DOR is defined as:
\begin{equation}
    DOR(\mathbf p,c)=(N_{(\mathbf p,c)} \times N_{(\bar {\mathbf p},\bar c)})/(N_{(\bar {\mathbf p},c)} \times N_{(\mathbf p,\bar c)}),
    \label{eq:dor}
\end{equation}
where $N_{(\mathbf p,c)}$ denotes the number of samples whose largest similarity to $\mathbf p$ is larger than 0.5 and the diagnosis is $c$, the number of samples is denoted as $N_{(\mathbf p,\bar c)}$ if the diagnosis is not $c$,  $N_{(\bar {\mathbf p}, c)}$ if the largest similarity is lower than 0.5, and $N_{(\bar {\mathbf p},\bar c)}$ if neither condition are fulfilled. 
95\% confidence interval is calculated as $DOR(\mathbf p,c) \pm 1.96 \times \sqrt{1/N_{(\mathbf p,c)}+1/N_{(\mathbf p,\bar c)}+1/N_{(\bar {\mathbf p},c)}+1/N_{(\bar {\mathbf p},\bar c)}}$. 
Prototypes with higher DOR values have a higher correlation with disease. We intend to increase the weights of prototypes that show high correlation with IK and to decrease the weights of prototypes that spuriously correlated to not-IK diseases,  where the refinement weights $\mathbf{w}_{ij}$ are calculated as follows:

\begin{equation}
   \text{\textit{global adjustment}:  }\hat{\mathbf{w}_{ij}} \leftarrow \frac{log(DOR(\mathbf P_{ij},i))}{\sum_{c=0 ,i \neq c}^{C}log(DOR(\mathbf P_{ij},c))} \mathbf{w}_{ij},
   \label{global_finetune}
\end{equation}
where $C$ is the number of all diseases.
Next, we will introduce how ophthalmologists and KGDM collaborate via sample-level interpretation during inferring. To incorporate the ophthalmologists' knowledge into the inferring phase, for each test image $X_0$, the adjusted KGDM provides an intermediate result of top $N$ prototypes ordered by weighted similarity as sample-level interpretation. Each of the prototypes with its critical region scaled from similarity map $S^{X}_{ij}$ is shown to physicians. The ophthalmologists are asked to visually compare the proposed critical regions and representative images$\mathbb{I}_{\mathbf P_{ij}}$, and give their subjective judgment on whether proposed diagnostic patterns are similar. After the audit of similarities, the ophthalmologists are asked to encode their assessment into a $N$ dimensional binary vector $\mathbf{d} \in \{0,1\}^{N}$. Then we can obtain a human intervened weighted sum of similarity $\hat{S_i}$ for disease $i$ as follows:

\begin{equation}
    \text{\textit{local discarding}:  }\hat{\tau_i} = \sum_{j=1}^{N} \mathbf{d}^j \hat{\mathbf{w}_{ij}} s^{X_0}_{ij},
\end{equation}
then the predicted probability of subclass $i$ is 
\begin{equation}
 p(i \mid X)=e^{\hat{\tau_i}} / \sum_{j=0}^{k} e^{\hat{\tau_j}} 
 \label{eq: prob}
\end{equation}

\noindent \textbf{Hyper-parameters and training details}
The KGDM model has been trained on the SRRT image dataset where the original image was captured with size 1024x768 or 2048x1536. During training, these images are resized to 384 according to the shorter edge for data augmentation. The augmentation includes random rotation, brightness jittering, and horizontal flipping. Then images with size $336x336$ are randomly cropped from augmented images. During testing and inference, images are directly resized and cropped $336x336$ from central.
To minimize the loss function, parameters are updated in $\{F(\cdot),g,\mathbf P, \mathbf w\}$ via back-propagation with the AdamW optimizer. For the initialization of $F(\cdot)$, we use pre-trained conventional neural networks as a backbone (e.g., \textit{ImageNet1K} of ResNet-50 \cite{He2016} provided by Pytorch \cite{Paszke2019
}). The prototype vectors are initialized from $m=10$ randomly selected example images of $k=4$ diseases, including IK, BK, FK, and HSK. The weights of the classification layer are initially set to 1.

For training hyperparameters, the batch size is set to 64, $\lambda_1$ is $10^{-3}$, and $\lambda_2$ to $10^{-3}$. $\lambda_{kl}$ is incremented from 0 to 1.
The learning rate is $5 \times 10^{-5}$ for $F(\cdot)$, $10^{-4}$ for $g$, and $10^{-2}$ for $\mathbf P$ and $\mathbf w$. betas of AdamW are 0.9 and 0.99. The number of epochs is set to 75 and the best AUC score is selected on the validation split of each fold. The backbone was first frozen for the first 10 epochs following the previous work \cite{Kim2021,Chen2019}, and then the backbone and prototype layers are trained jointly in the rest epochs. All training processes were conducted on 4 pieces of Nvidia GeForce RTX 3090 GPU.

\subsection*{Evaluate and Metrics}
\textbf{ Human-XAI collaboration test}
\label{sec:human_ai}
12 ophthalmologists from SRRSH participated in the study as independent and blinded test readers. We separated the testers into two groups, the 6 clinicians in the group `Junior ophthalmologists (Junior)' were residents with professional experience of fewer than 5 years, while the other 6 doctors in the group  `Senior ophthalmologists (Senior)' had at least Fellow or higher academic titles, whose professional experience were more than 5 years. No ophthalmologists involved in the ground truth creation participated as readers. 
The  Human-XAI collaboration test followed the two-step protocol. In the first step, an ophthalmologist conducted an image-only diagnosis. Images of the four categories of IK (AK, BK, FK, and HSK), other corneal diseases, and corneas without significant abnormalities, were presented to the ophthalmologist, who made a diagnostic decision from these six choices for each image through manual examination, the uncertainty of each decision was judged at the same time. Then in the second step, the same 140 images were presented to the tester after randomization. The ophthalmologist collaborated with the KGDM, which provided him with the sample-level interpretation of the prediction of KGDM, the contribution of representative regions to KGDM prediction, and prototypes' class-level interpretation information, including visualization of $\mathbb I_{\mathbf P}$. The tester was supposed to remake a diagnosis decision after receiving all this information and judge it with his professional knowledge and clinical experience.
Before the test, the instructions of the Human-XAI collaboration test platform were provided to the ophthalmologist, and adequate explanations were offered to make sure the tester fully understand the design and operation principle of our KGDM. All ophthalmologists performed this procedure independently and without time limitations.

%% file: supplementary.tex
\appendix
\newpage

\setcounter{table}{0}
\renewcommand{\thetable}{S\arabic{table}}
\setcounter{figure}{0}
\renewcommand{\thefigure}{S\arabic{figure}}

\section*{Supplementary Materials}
\subsection*{List of supplementary materials}
Table S1 to S4
Figures S1 to S4


\begin{figure*}[ht]
    \includegraphics[width=\linewidth]{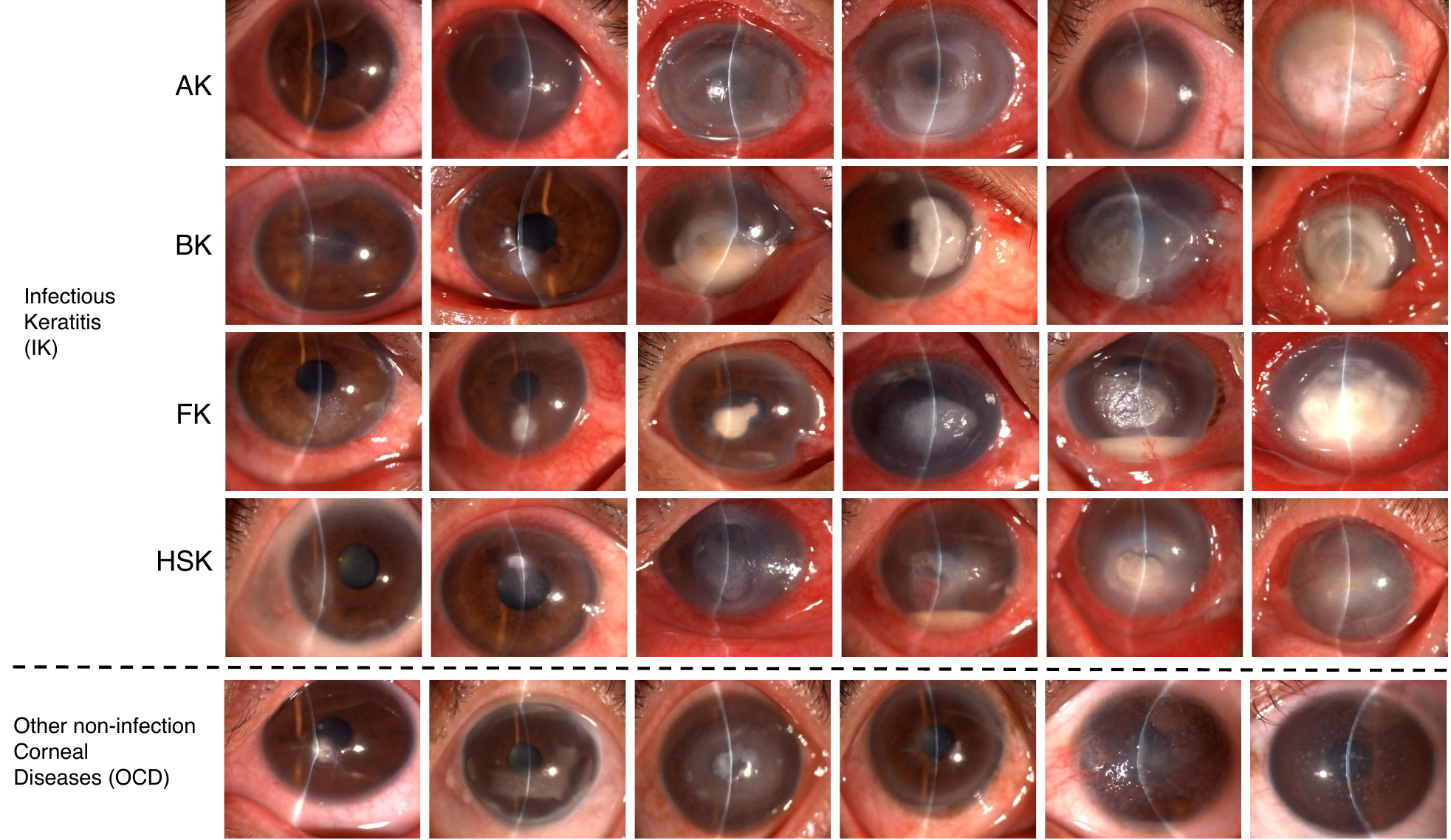}
    \caption{\textbf{Representative images of infectious keratitis and other corneal diseases} 
    The figure shows representative slit lamp microscopic images of AK, BK, FK, HSK and other corneal diseases. The first four rows of images respectively present the characteristics of the aforementioned four types of infectious keratitis at different stages, while the last row of images displays the features of other categories of corneal diseases excluding infectious keratitis. }
    \label{fig:exampleIK}
\end{figure*}

\begin{figure*}[ht]
    \centering
    \includegraphics[width=\linewidth]{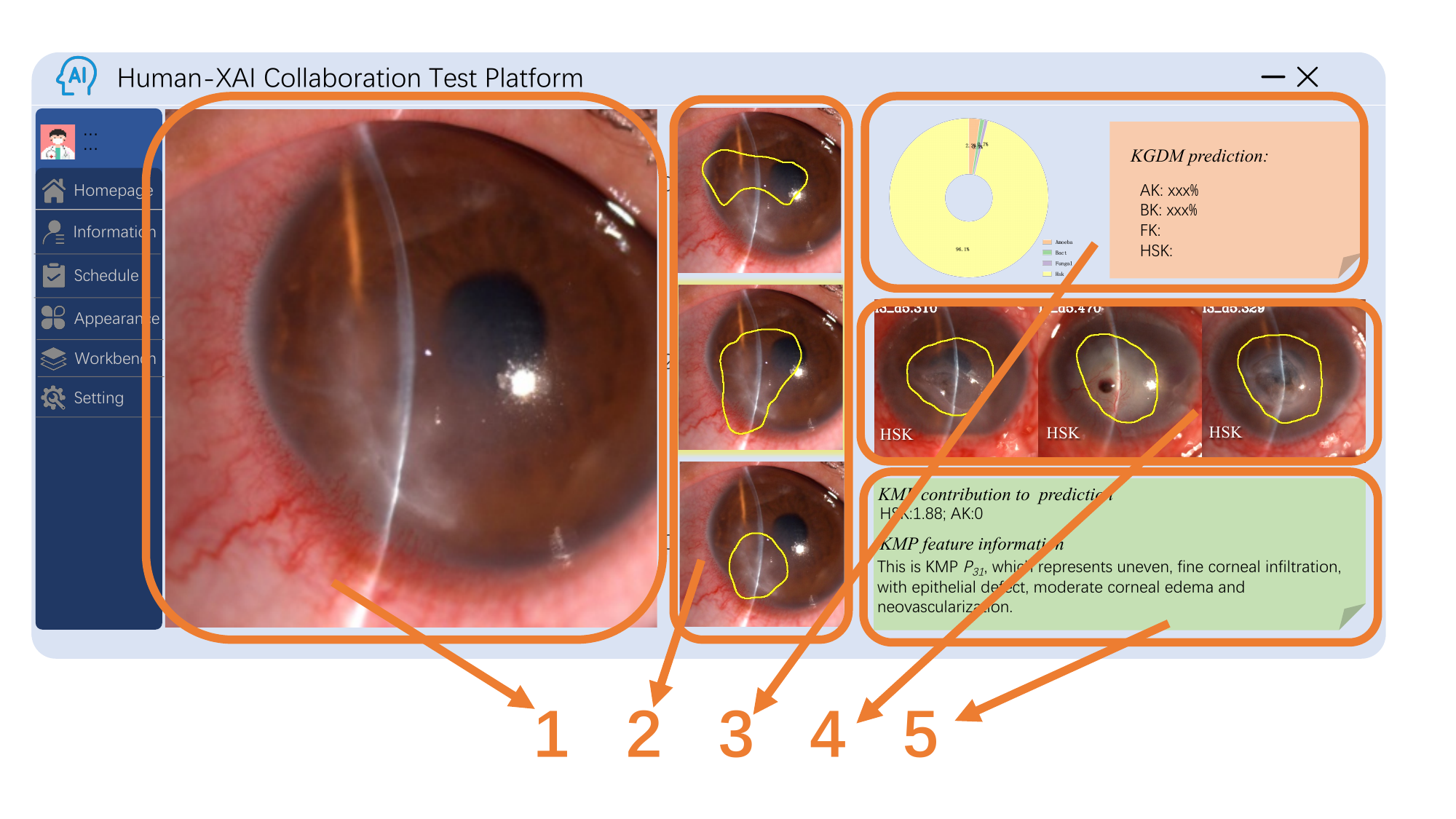}
    \caption{\textbf{Human-XAI collaboration interface} 
    The test image is presented in Region 1. The prediction of KGDM for this image can be found in Region 3, the pie chart indicates the probability for each classification. In Region 2, the 3 KMPs with the highest contributions were visualized, the user can switch between the three for more detailed information on them (as shown in Region 4 and Region 5). For each KMP, the similar regions in the training dataset are presented in Region 4 ensuring an intuitive understanding of the KMP feature for the tester. Region 5 exhibits the quantitative value of contribution for each classification and the description of the corresponding KMP. In combination with the information above, the physician reconsiders the diagnosis of each image and makes a final decision from the 6 options.}
    \label{fig:hmc}
\end{figure*}

\begin{figure*}
    \centering
    \includegraphics[width=\linewidth]{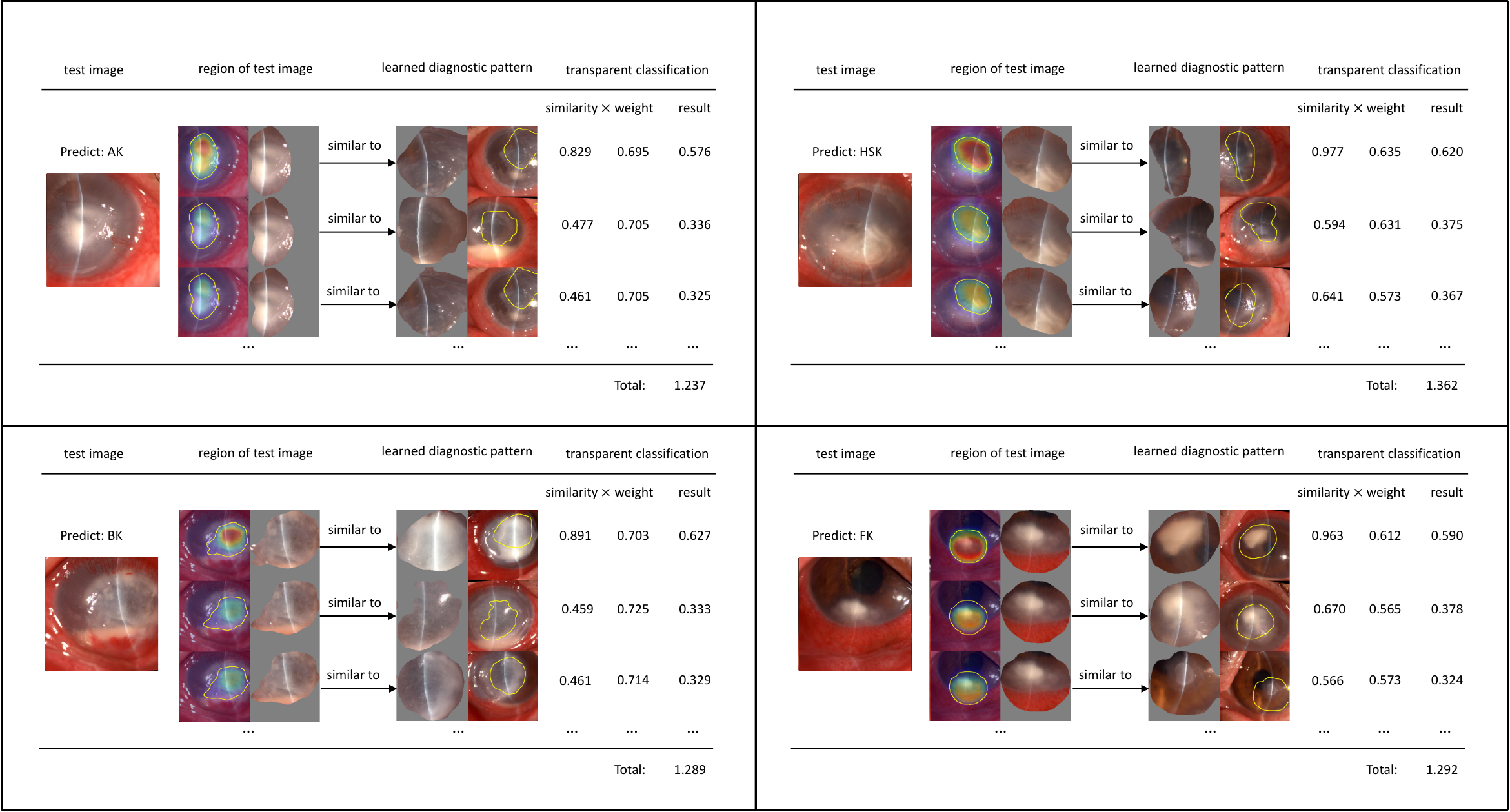}
    \caption{\textbf{Local interpretation examples of KGDM} 
        The local interpretation results for AK, BK, FK, and HSK effectively illustrate how KGDM identifies diagnostic keratitis manifestation patterns in these images, as visually delineated by yellow lines. Heatmaps further accentuate this information, while historical cases from the training set provide supportive evidence for these patterns and their respective quantitative contributions. These insightful findings aid clinicians in comprehending how AI classifies each sample.}
    \label{fig:ext_interp}
\end{figure*}

\begin{table*}[!ht]
    \centering
    \resizebox{\textwidth}{!}{
    \begin{tabular}{lllll}
    \hline
        image & label & URL & source & filename \\ \hline
        1 & amoeba & eye-infect1-img.jpg (318×217) (2020mag.com) & magzine report & eye-infect1-img.jpg \\ 
        6 & amoeba & acanthamoeba-LRG.jpg (2048×1349) (uiowa.edu) & the university of lowa  & acanthamoeba-LRG.jpg \\ 
        7 & amoeba & 016\_rp0420\_cas.jpg (847×937) (reviewofophthalmology.com) & article & acanthamoeba-LRG.jpg \\ 
        8 & amoeba & Acantha.wiki\_.tc\_.jpg (320×270) (choroida.com) & case report & acanthamoeba-LRG.jpg \\ 
        9 & amoeba & 15911050004498272775ed655e818537.png (1138×800) (eophtha.com) & article & acanthamoeba-LRG.jpg \\ 
        12 & amoeba & OPHTH2021-\_5\_\_Image-2.jpg (1920×1080) (ctfassets.net) & the Cornea Service at New York Eye and Ear Infirmary of Mount Sinai & OPHTH2021-\_5\_\_Image-2.jpg \\ 
        13 & amoeba & [Figure, Acanthamoeba keratitis. Image courtesy S Bhimji MD] - StatPearls - NCBI Bookshelf (nih.gov) & national library of medicine & acanthamoeba\_\_keratitis.jpg \\ 
        14 & amoeba & toptips-fig-1.png (480×475) (eyenews.uk.com) & news report & toptips-fig-1.png  \\ 
        15 & amoeba & Hyperreflective tissue in corneal ulceration of severe Acanthamoeba... | Download Scientific Diagram (researchgate.net) & article & Hyperreflective-tissue-in-corneal-ulceration-of-severe-Acanthamoeba-keratitis-cases.png \\ 
        16 & amoeba & 2.png (418×281) (clinicasnovovision.com) & web & 2.png  \\ 
        17 & amoeba & IndianJOphthalmol\_2017\_65\_11\_1079\_218084\_f7.jpg (809×559) (ijo.in) & article & IndianJOphthalmol\_2017\_65\_11\_1079\_218084\_f7.jpg \\ 
        18 & amoeba & omd\_nov\_s2301.jpg (392×216) (ophthalmologymanagement.com) & article & omd\_nov\_s2301.jpg \\ 
        20 & amoeba & 127-124905-contact-lenses-blindness-acanthamoeba-3.jpeg (768×511) (al-ain.com) & ~ & 127-124905-contact-lenses-blindness-acanthamoeba-3.jpeg \\ 
        21 & amoeba & a54292\_23165f7d08d94955999b624362b5532e\~mv2\_d\_4928\_3264\_s\_4\_2.webp (3376×2236) (wixstatic.com) & web & a54292\_23165f7d08d94955999b624362b5532e\~mv2\_d\_4928\_3264\_s\_4\_2.webp \\ 
        1 & bact & https://www.ncbi.nlm.nih.gov/pmc/articles/PMC3342793/bin/i1552-5783-53-4-1787-f01.jpg & article & i1552-5783-53-4-1787-f01A.jpg \\ 
        2 & bact & https://www.ncbi.nlm.nih.gov/pmc/articles/PMC3342793/bin/i1552-5783-53-4-1787-f01.jpg & article & i1552-5783-53-4-1787-f01B.jpg \\ 
        6 & bact & htmlconvd-qd6DgS87x1.jpg (639×936) (studfile.net) & article & htmlconvd-qd6DgS87x1.jpg \\ 
        7 & bact & 3-s2.0-B9780128127353005380-f13-03-9780128127353.jpg (355×236) (els-cdn.com) & Encyclopedia of Pharmacy Practice and Clinical Pharmacy & 3-s2.0-B9780128127353005380-f13-03-9780128127353.jpg \\ 
        9 & bact & https://www.ophthalmologytraining.com/images/Photos/Bacterial\%20Keratitis.jpg & web & Bacterial Keratitis.jpg \\ 
        12 & bact & image02590.jpeg (413×276) (oncohemakey.com) & web & image02590.jpeg \\ 
        15 & bact & 9781451180398\_fig7-1d\_FIG.jpg (384×258) (entokey.com) & web & 9781451180398\_fig7-1d\_FIG.jpg \\ 
        16 & bact & 321de5ec-8c4a-4a52-85a8-a7e6bda4dc5e.webp (300×229) (d31g6oeq0bzej7.cloudfront.net) & American Academy of Ophthalmology & 321de5ec-8c4a-4a52-85a8-a7e6bda4dc5e.webp \\ 
        17 & bact & microbial-keratitis-1.jpg (866×578) (plano.co) & web & microbial-keratitis-1.jpg \\ 
        18 & bact & 9781451180398\_fig7-1b\_FIG.jpg (384×255) (entokey.com) & web & 9781451180398\_fig7-1b\_FIG.jpg \\ 
        19 & bact & 3-s2.0-B9780128127353005380-f13-04-9780128127353.jpg (355×236) (els-cdn.com) & Encyclopedia of Pharmacy Practice and Clinical Pharmacy & 3-s2.0-B9780128127353005380-f13-04-9780128127353.jpg \\ 
        20 & bact & AJGP-08-2019-Focus-Lee-Microbial-Keratitis-Fig-2.jpg.aspx (760×427) (racgp.org.au) & article & AJGP-08-2019-Focus-Lee-Microbial-Keratitis-Fig-2A.jpg \\ 
        21 & bact & AJGP-08-2019-Focus-Lee-Microbial-Keratitis-Fig-2.jpg.aspx (760×427) (racgp.org.au) & article & AJGP-08-2019-Focus-Lee-Microbial-Keratitis-Fig-2B.jpg \\ 
        22 & bact & AJGP-08-2019-Focus-Lee-Microbial-Keratitis-Fig-2.jpg.aspx (760×427) (racgp.org.au) & article & AJGP-08-2019-Focus-Lee-Microbial-Keratitis-Fig-2C.jpg \\ 
        1 & fungal & https://markallen.blob.core.windows.net/launchpad/course-images/0e4be4a3-f4a4-43a2-8dd1-7afa6ce3b98c.jpg & web & 9781451180398\_fig7-1b\_FIG.jpg \\ 
        2 & fungal & The-right-eye-of-the-case-1-with-deep-corneal-ulcer-and-stromal-infiltration.jpg (600×436) (researchgate.net) & article & The-right-eye-of-the-case-1-with-deep-corneal-ulcer-and-stromal-infiltration.jpg \\ 
        3 & fungal & f007-003-9780702029837.jpg (387×329) (wp.com) & web & f007-003-9780702029837.jpg \\ 
        4 & fungal & 0adfc1a5-4644-4aa5-862b-4c5d3db9189d.webp (830×415) (d31g6oeq0bzej7.cloudfront.net) & article & 0adfc1a5-4644-4aa5-862b-4c5d3db9189d.webp  \\ 
        5 & fungal & 1a-fungal.jpg (600×450) (uiowa.edu) & the university of lowa  & 1a-fungal.jpg \\ 
        7 & fungal & Fungal\_ulcer.jpeg (604×480) (aao.org) & web & Fungal\_ulcer.jpeg \\ 
        8 & fungal & 1c-fungal.jpg (600×450) (uiowa.edu) & the university of lowa  & 1c-fungal.jpg \\ 
        9 & fungal & 2\_1465\_2.jpg (200×147) (reviewofoptometry.com) & web & 2\_1465\_2.jpg \\ 
        10 & fungal & 12886\_2015\_92\_Fig1\_HTML.gif (472×316) (springernature.com) & article & 12886\_2015\_92\_Fig1\_HTML.gif \\ 
        11 & fungal & 209759.fig.003a.jpg (600×433) (hindawi.com) & article & 12886\_2015\_92\_Fig1\_HTML.gif \\ 
        13 & fungal & fungal-keratitis-1.jpg (1125×769) (life-worldwide.org) & web & fungal-keratitis-1.jpg \\ 
        15 & fungal & CommunityAcquirInfect\_2015\_2\_4\_142\_172648\_f1.jpg (619×569) (caijournal.com) & case report & CommunityAcquirInfect\_2015\_2\_4\_142\_172648\_f1.jpg \\ 
        16 & fungal & fungal-keratitis-201208-LRG.jpg (1353×1160) (uiowa.edu) & the university of lowa  & fungal-keratitis-201208-LRG.jpg \\ 
        17 & fungal & 016\_RCCL0916\_BiofilmFormation-3.jpg (300×203) (reviewofcontactlenses.com) & web & 016\_RCCL0916\_BiofilmFormation-3.jpg \\ 
        18 & fungal & Figure 4 from A case of radial keratoneuritis in non-Acanthamoeba keratitis | Semantic Scholar & article & 3-Figure4-1.png \\ 
        19 & fungal & microbial-keratitis-2.jpg (976×651) (plano.co) & web & microbial-keratitis-2.jpg  \\ 
        20 & fungal & Figure - PMC (nih.gov) & article & nje-07-685-g004.jpg \\ 
        21 & fungal & image.axd (609×415) (aao.org) & web & image.axd \\ 
        1 & HSK & 1559831556-0619\_Cornea\_Fig3.png (608×416) (imgix.net) & web & 1559831556-0619\_Cornea\_Fig3.png \\ 
        2 & HSK & Disciform-LRG.jpg (1900×1266) (uiowa.edu) & the university of lowa  & Disciform-LRG.jpg \\ 
        3 & HSK & Disciform-SS-LRG.jpg (2048×1365) (uiowa.edu) & the university of lowa  & Disciform-SS-LRG.jpg \\ 
        7 & HSK & 026\_RCCL1120\_F3\_CE\_Leon\_3.jpg (1200×907) (revieweducationgroup.com) & web & 026\_RCCL1120\_F3\_CE\_Leon\_3.jpg \\ 
        8 & HSK & 029\_RCCL1120\_F3\_CE\_Leon\_6.jpg (1200×898) (revieweducationgroup.com) & web & 029\_RCCL1120\_F3\_CE\_Leon\_6.jpg \\ 
        12 & HSK & 027\_RCCL1120\_F3\_CE\_Leon\_4.jpg (1200×907) (revieweducationgroup.com) & web & 027\_RCCL1120\_F3\_CE\_Leon\_4.jpg  \\ 
        14 & HSK & 010\_RCCL1117\_CC-1.jpg (480×286) (reviewofcontactlenses.com) & web & 010\_RCCL1117\_CC-1.jpg \\ 
        15 & HSK & JEK\_Fig2.jpg (373×313) (aao.org) & web & JEK\_Fig2.jpg \\ 
        18 & HSK & https://ars.els-cdn.com/content/image/3-s2.0-B9780128187319001099-f00109-04-9780128187319.jpg & article & 3-s2.0-B9780128187319001099-f00109-04-9780128187319.jpg \\ 
        19 & HSK & IKWOdYfwOhORr2c2o0XMAMnxXDoHWlFjKUDazDla.jpeg (600×348) & web & IKWOdYfwOhORr2c2o0XMAMnxXDoHWlFjKUDazDla.jpeg  \\ 
        20 & HSK & HSV-dendritic-ulcer.jpg.webp (580×435) (litfl.com) & web & HSV-dendritic-ulcer.jpg \\ 
        21 & HSK & 01\_nov\_2016\_figure\_1g90.jpg (533×400) (dovepress.com) & article & 01\_nov\_2016\_figure\_1g90.jpg \\ 
        22 & HSK & https://emedicine.medscape.com/ article/1194268-treatment & article & Figure4.jpg \\ 
        23 & HSK & ophthal-image-1.jpg (389×263) (mivision.com.au) & web & ophthal-image-1.jpg \\ 
        24 & HSK & HSV-disciform-keratitis.jpg (800×600) (eyerounds.org) & the university of lowa  & HSV-disciform-keratitis.jpg \\ 
        25 & HSK & chto-takoe-keratit-3-768x683.jpg (768×683) (microbak.ru) & web & chto-takoe-keratit-3-768x683.jpg \\ \hline
    \end{tabular}
    }
    \caption{\textbf{Reference of each sample in PUB dataset.} This table presented the labels, URL, source and filenames of each image collected from the Internet, which made up the online available publication testing dataset. }
    \label{tab:pub_download}
\end{table*}

\begin{table}[ht]
    \centering
    \caption{\textbf{Ablation study of KGDM with ResNet50 as backbone.}}
    \begin{threeparttable}
        \resizebox{\textwidth}{!}{
            \begin{tabular}{lcccc}
                \rowcolor{Dark}
                \hline 
                \multicolumn{1}{c}{\text { Metrics }} & \text { AUROC (95\% CI) } & \text { Cohen's Kappa (95\% CI) } & \text { Sensitivity (95\% CI) } & \text { PPV (95\% CI) } \\
                \hline
                \rowcolor{Gray}
                \multicolumn{5}{l}{\textbf {(a) Internal Retrospective Dataset: SRRT }}\\
                 ResNet50 & 0.899(0.887-0.909) &  0.648(0.631-0.663) &  68.84\%(67.20\%,70.50\%) &  71.37\%(69.90\%,72.76\%) \\

                 +prototype layer  &     0.889(0.881-0.898) &  0.621(0.605-0.637) &  65.73\%(64.12\%,67.39\%) &  69.72\%(67.58\%,71.63\%) \\
                 +$\mathcal{L}_{prior}$  &  0.930(0.918-0.940) &  0.773(0.745-0.797) &  77.08\%(74.25\%,79.68\%) &  82.58\%(79.76\%,85.05\%) \\
                 +prototype layer($\mathcal{L}_{prior})$  &  0.878(0.864-0.891) &  0.615(0.591-0.638) &  65.48\%(63.52\%,67.48\%) &  69.27\%(66.55\%,72.05\%) \\
                 KGDM  &  0.915(0.903-0.928) &  0.740(0.714-0.772) &  74.54\%(72.08\%,77.40\%) &  80.54\%(77.85\%,83.24\%) \\
                
                 \rowcolor{Gray}
                \multicolumn{5}{l}{\textbf {(b) External Retrospective Dataset: XS }}\\
                ResNet50 & 0.726(0.713-0.738) &   0.089(0.079-0.099) &  43.99\%(42.08\%,45.88\%) &  24.99\%(22.67\%,27.02\%) \\
                +prototype layer &  0.735(0.712-0.763) &   0.170(0.159-0.180) &  60.76\%(57.89\%,62.90\%) &  32.35\%(27.98\%,36.57\%) \\
                $\mathcal{L}_{prior}$ &  0.775(0.751-0.794) &  0.192(0.174-0.208) &  65.63\%(62.41\%,68.61\%) &  40.84\%(36.14\%,45.12\%) \\
                +prototype layer($\mathcal{L}_{prior}$) &  0.726(0.707-0.743) &  0.174(0.163-0.184) &  62.09\%(59.90\%,64.19\%) &  36.33\%(31.47\%,40.52\%) \\
                 KGDM &  \textbf{0.770(0.723-0.817)} & \textbf{ 0.411(0.313-0.509)} &  \textbf{52.83\%(44.51\%-61.15\%)} &  \textbf{61.07\%(46.71\%-75.43\%)} \\
                
                 \rowcolor{Gray}
                \multicolumn{5}{l}{\textbf {(c) External Retrospective Dataset: PUB } }\\
                ResNet50 &  0.521(0.420-0.623) &  0.033(-0.182-0.249) &  26.14\%(10.89\%-41.39\%) &  27.68\%(8.31\%-47.05\%) \\
                +prototype layer & 0.582(0.555-0.606) &   0.098(0.059-0.135) &  31.47\%(28.33\%,34.74\%) &  42.85\%(35.43\%,49.79\%)  \\
                 $\mathcal{L}_{prior}$ &  0.624(0.577-0.671) & \textbf{ 0.183(0.121-0.247)} &  35.55\%(30.84\%,40.58\%) &  39.05\%(28.22\%,50.95\%) \\
                +prototype layer($\mathcal{L}_{prior}$) &  0.543(0.520-0.568) &  0.098(0.053-0.145) &  31.10\%(27.63\%,34.84\%) &  39.40\%(31.46\%,46.82\%) \\
                 KGDM &  \textbf{0.638(0.555-0.721)} &  \textbf{0.181(0.035-0.328)} & \textbf{ 37.69\%(25.89\%-49.48\%)} & \textbf{ 55.38\%(42.35\%-68.42\%)} \\
                
                 \rowcolor{Gray}
                \multicolumn{5}{l}{\textbf {(d) Internal  Prospective Dataset: SRRPV }}\\

                ResNet50 & 0.759(0.674-0.843) & 0.348(0.154-0.542) & 48.10\%(32.83\%-63.37\%) & 51.44\%(38.74\%-64.15\%) \\
                +prototype layer &  0.782(0.753-0.806) &   0.440(0.401-0.480) &  57.06\%(54.38\%,60.11\%) &  60.07\%(56.80\%,63.63\%) \\
                 $\mathcal{L}_{prior}$ &  0.820(0.799-0.839) &  0.507(0.435-0.575) &  60.08\%(55.34\%,64.47\%) &  68.50\%(61.62\%,74.46\%) \\
                 +prototype layer($\mathcal{L}_{prior}$) &  0.789(0.769-0.809) &  0.456(0.419-0.487) &  57.98\%(55.71\%,60.29\%) &  63.15\%(58.94\%,67.33\%) \\
                 KGDM  &   \textbf{0.836(0.811-0.861)} &   \textbf{0.558(0.521-0.596)} &   \textbf{62.65\%(59.88\%-65.41\%)} &   \textbf{70.03\%(65.20\%-74.86\%)} \\
                \hline
            \end{tabular}
            }
    
    \begin{tablenotes}
        \footnotesize
        \item[1] F1 score combines the precision and recall of a classifier into a single metric by taking their harmonic mean.
    \end{tablenotes}
    \end{threeparttable}
    \label{tab:ablation}
\end{table}

\begin{table}[ht]
    \centering
    \caption{\textbf{Performance of KGDM on SRRPV dataset under different uncertainty threshold.}}
    \begin{threeparttable}
        \resizebox{\textwidth}{!}{
            \begin{tabular}{lccccc}
                \rowcolor{Dark}
                \hline 
                \multicolumn{1}{c}{\text { Confidence Level }} & \text { AUROC (95\% CI) } & \text { Cohen's Kappa (95\% CI) } & \text { Sensitivity (95\% CI)} & \text { PPV (95\% CI) }   \\
                \hline
                0.1 &  0.858(0.803-0.918) &  0.722(0.650-0.814) &  66.69\%(57.50\%,77.71\%) &  69.83\%(55.76\%,83.28\%) \\
                0.2 &  0.846(0.788-0.898) &  0.673(0.597-0.742) &  65.84\%(59.89\%,73.03\%) &  75.57\%(65.58\%,84.66\%) \\
                0.3 &  0.847(0.807-0.886) &  0.648(0.586-0.703) &  65.26\%(61.03\%,69.71\%) &  76.59\%(67.08\%,84.83\%) \\
                0.4 &  0.841(0.807-0.879) &  0.638(0.567-0.694) &  66.00\%(61.15\%,70.44\%) &  79.59\%(71.74\%,85.03\%) \\
                0.5 &  0.836(0.809-0.866) &  0.631(0.561-0.691) &  66.09\%(61.67\%,70.51\%) &  78.34\%(70.90\%,84.14\%) \\
                0.6 &  0.833(0.806-0.863) &  0.601(0.530-0.664) &  64.64\%(59.83\%,69.75\%) &  75.18\%(67.98\%,81.48\%) \\
                0.7* &  0.829(0.805-0.853) &  0.572(0.499-0.631) &  63.48\%(58.44\%,68.01\%) &  73.10\%(66.80\%,78.76\%) \\
                0.8 &  0.820(0.799-0.839) &  0.507(0.435-0.575) &  60.08\%(55.34\%,64.47\%) &  68.50\%(61.62\%,74.46\%) \\
                0.9 &  0.806(0.785-0.823) &  0.441(0.370-0.503) &  57.07\%(52.44\%,61.45\%) &  63.31\%(56.58\%,68.77\%) \\
                1.0 &  0.806(0.785-0.823) &  0.441(0.370-0.503) &  57.07\%(52.44\%,61.45\%) &  63.31\%(56.58\%,68.77\%) \\
                \hline
            \end{tabular}
            }
        \begin{tablenotes}
        \footnotesize
        \item[*] empirical confidence level 0.7 was chosen. Smaller confidence level leads to less misclassifying non-IK diseases as IK, but too small might leads high false-negative.
            \end{tablenotes}
        
    \end{threeparttable}

\label{tab:performance_app}
\end{table}

\begin{table}
    \centering
    \caption{Retrieve accuracy of AI-based biomarkers. We reported the top-1, top-3 and top-5 accuracy of each biomarker in all activate test sample in SRRPV.}
    \begin{tabular}{lrrr}
        \toprule
         disease &  acc@1 &  acc@3 &  acc@5 \\
        \midrule
        Amoeba &      0.78 & 0.92 &    0.98 \\
        Bact &      0.41 & 0.77 &    0.88 \\
        Fungal &       0.92 & 0.95 &    1.00\\
        HSK &       0.87 & 0.98 &    1.00 \\
        \bottomrule
        \end{tabular}
    \label{tab:retrieved_samples_accuracy}
\end{table}

\begin{table}[]
    \centering
    \caption{Diagnostic odds ratio of AI-based biomarkers}
    \begin{tabular}{llrlr}
        \toprule
        name &       disease &  odd\_ratio &              CI95 &  pvalue \\
        \midrule
          P0 &        Amoeba &      28.60 & (24.0708-33.9741) &    0.00 \\
          P1 &        Amoeba &      21.01 & (12.4466-35.4559) &    0.00 \\
          P2 &        Amoeba &      43.51 & (37.7746-50.1081) &    0.00 \\
          P4 &        Amoeba &      25.96 & (21.1432-31.8793) &    0.00 \\
          P7 &        Amoeba &      17.34 & (15.0962-19.9233) &    0.00 \\
          P8 &        Amoeba &      48.11 & (41.7572-55.4349) &    0.00 \\
         P11 &          Bact &       5.52 &  (0.5001-60.9029) &    0.23 \\
         P15 &          Bact &       5.55 &  (2.4867-12.3699) &    0.00 \\
         P19 &          Bact &       5.40 &   (3.9548-7.3619) &    0.00 \\
         P22 &        Fungal &       7.27 &   (6.4480-8.1864) &    0.00 \\
         P28 &        Fungal &      13.40 & (11.6998-15.3558) &    0.00 \\
         P31 &           Hsk &       7.80 &   (6.3628-9.5732) &    0.00 \\
         P33 &           Hsk &       7.97 &   (6.4965-9.7737) &    0.00 \\
         P35 &           Hsk &       7.33 &   (5.9756-8.9924) &    0.00 \\
         P39 &           Hsk &       7.93 &   (6.4687-9.7320) &    0.00 \\
        \bottomrule
        \end{tabular}
    \label{tab:dor_scores}
\end{table}